# Source codes in human communication

Michael Ramscar

Department of Linguistics

University of Tübingen

**Abstract**

Although information theoretic characterizations of human communication have become increasingly popular in linguistics, to date they have largely involved grafting probabilistic constructs onto older ideas about grammar. Similarities between human and digital communication have been strongly emphasized, and differences largely ignored. However, some of these differences matter: communication systems are based on predefined codes shared by every sender/receiver, whereas the distributions of words in natural languages guarantee that no speaker/hearer ever has access to an entire linguistic code, which seemingly undermines the idea that natural languages are probabilistic *systems* in any meaningful sense. This paper describes how the distributional properties of languages meet the various challenges arising from the differences between information systems and natural languages, along with the very different view of human communication these properties suggest. To illustrate, it presents an account of the semantic function of personal names (a traditional linguistic stumbling block), describing the non-compositional, discriminative communicative process supported by the statistical and syntactic properties of names in the world's languages. It also describes how this account extends to the rest of human communication, and describes how its predictions are supported by analyses the distribution of English nouns and verbs.

## 1. Human communication and compositionality

For as long as humans have contemplated their unique communicative abilities, the way that language works has overwhelmingly been thought about in terms of 'conduit' metaphors (Reddy, 1979). It has been supposed that utterances somehow 'contain' (or point to) their meanings, such that a speaker packs (or encodes) the meaning of a message into words, which is then unpacked (or extracted, or decoded) by a listener. Accordingly, natural languages have been thought of as comprising various kinds of discrete 'units' or packages that somehow contribute to the conveyance (or signification) of meanings. These units/packages are usually conceived of at various levels of description (Ramscar & Port, 2016): *phonemes* are roughly the acoustic/psychological version of letters; *morphemes*, are 'atomic' units of meaning that cannot be subdivided (such that the word *rain-bow-s*, = 3 morphemes); *words*, which can be either mono- or multi-morphemic and which can be easier or harder to identify (e.g., *tree vs dunno*); and *sentences*, which are larger packages comprising multiple words / morphemes. These various levels of description are unified by the <u>principle of compositionality</u>, which holds that the meaning conveyed by an utterance is the sum of the meanings of the morphemes and words it comprises.

However, attempts to make concrete sense of the idea of discrete linguistic units have encountered a familiar range of critical problems at every one of these levels of description. Despite their intuitive appeal, phonemes fail to capture many of the acoustic properties essential to spoken communication, and critically these basic sound units – that are supposed to 'spell out' all of the other elements involved in compositional processes – are routinely underdetermined in speech (i.e., they are impossible to identify from acoustic information alone), such that their 'presence' in a speech signal can only be inferred from context (Port & Leary, 2005; Ramscar & Port, 2016). The idea of morphemes as basic units linking sound to meaning has been assailed on a number of fronts: by linguistic analyses of their putative functions (which often seem to be meaningless), and again by the discovery of numerous effects of context on supposedly fixed units (Blevins, 2006; 2016). Meanwhile psychological research has, after half a century of motivated effort, singularly failed to find support for the existence of units of meanings (Ramscar & Port, 2015), while philosophical analyses that have long suggested that the whole idea of meaning units is fundamentally misguided (Wittgenstein, 1953; Quine, 1960). Accordingly, and perhaps unsurprisingly, the various problems that have arisen in trying to explain words and morphemes have resurfaced as researchers have tried to explain how 'larger meanings' are supposed to be systematically computed as smaller elements of meaning combine in sentences (Culicover, 1999).

The theoretical attraction of compositionality – and the recurrent problems it has posed for linguists – can be summarized as follows: By presupposing the existence of a context independent inventory of elements (phonemes, morphemes, words, etc.) out of which larger meanings are built, compositionality offers up the theoretical promise of reversing the process, enabling linguists to show how larger meanings decompose into their elements. Unfortunately however, when the results of the many good-faith attempts to identify context independent 'units' are analyzed with this in mind, what soon becomes apparent is that the understanding of natural languages is influenced by context at *every* possible level of description. Which is to say that the results of the considerable body of research aimed at identifying context independent sound / meaning elements point inexorably in a single direction: that, empirically, the intuitions that have led theorists to suppose that human communication relies on the composition and decomposition of basic elements or units (or whatever) are likely to be wrong.

If one cares about building a scientific account of human communication, this is a problem. Virtually all of linguistic theory is predicated on the idea of compositionality and compositional units, such that the current literature abounds with 'findings' that only make scientific sense if ones invokes some combination of phonemes, morphemes, words or sentences (and invokes them as ontologically important linguistic / psychological elements, not just descriptively useful constructs). Since the evidence suggests that the various units proposed by linguists have no real, ontological status in human communication, it thus follows that the theoretical state of the art in language science may well be less a body of knowledge than a bubble, an enterprise founded on promissory notes that can never be cashed in.[1]

From this perspective, the resurgence of interest in information theoretical approaches to linguistics (e.g. e.g. Pereira, 2000; Hale, 2001; van Son, & Pols, 2003; Aylett, & Turk, 2004; Levy, 2008; Milin, Đurđević, & del Prado Martín, 2009; Milin, Kuperman, Kostic, & Baayen, 2009; Jaeger, 2010; Piantadosi, Tily, & Gibson, 2011, 2012; Ackerman & Malouf, 2013; Futrell, Mahowald, & Gibson, 2015; Westbury, Shaoul, Moroschan, & Ramscar, 2016; Dye, Milin, Futrell & Ramscar, 2017; Meylan & Griffiths, 2017; Dye, Milin, Futrell & Ramscar, 2018) seems at first blush to offer a positive response to the discovery that context exerts a pervasive influence on the form and content of human communication, since information theory is explicitly concerned with quantifying context, and structuring it to optimize signaling (Hartley, 1928; Shannon, 1948). However, even linguists who embrace the idea that language use is fundamentally contextual and probabilistic (see also e.g., Christiansen & Chater, 2008; Evans & Levinson, 2009), still tend overwhelming to adhere to a compositional account of language. Moreover, because even probabilistic approaches to linguistic theory have tended to graft probabilistic constructs onto older ideas about the nature of grammar (rather than using them as the basis for a fundamental rethink of the nature of linguistic processing), 'information theoretic' approaches in linguistics have largely tended to emphasize similarities between human and digital communication, while ignoring the many important differences between them.

---

[1] One approach to this problem is to simply reject out of hand the empirical data that makes the identification of units seem such a recalcitrant problem. Since it has been often been argued that language cannot be learned from 'surface forms' (i.e, empirical data), some researchers have taken this line of thought to its logical conclusion and argued that since linguistic units – including discrete word meanings – appear to be under-determined in experience, they too must be innate (Chomsky, 2000). While these arguments are largely unfalsifiable, they are still undermined by results showing how supposedly unlearnable aspects of natural language 'syntax' are learned (Ramscar et al, 2007, 2013a), how the learning of communicative codes conforms to the predictions of standard learning mechanisms (Ramscar et al, 2010, 2013b,c) and by advances in machine learning, which have show how surprisingly accurate speech recognition can be learned from empirical data (Hannun et al. 2014).

This is a problem, because although they are barely recognized in the literature, some of the apparent differences between natural languages and information theoretic communication systems seem to undermine not only information theoretic approaches to human communication, but also *any* probabilistic theory of language, because they appear to undermine the idea that natural languages are shared *systems* in any meaningful sense:

1. Shannon's (1948) theory of communication is a <u>system</u> solution to the problem of signaling over a noisy channel (Mackay, 2003). That is, digital communication systems are based on predefined *source* and *channel codes*,[2] which define probabilistic models that are shared by every sender/receiver (Shannon, 1948). However, the distribution of words in natural languages, and the fact that people are required to *learn* the language they use, appear to statistically guarantee that <u>no</u> speaker / hearer ever has access to an entire communicative code. Individual communicative codes thus appear to be always incomplete, such that in practice, every speaker constantly encounters unattested forms – i.e., aspects of the code that were hitherto outside their experience (Ramscar, Hendrix, Love & Baayen, 2013; Ramscar, Hendrix, Shaoul, Milin & Baayen, 2014; Keuleers, Stevens, Mandera, and Brysbaert, 2015; Blevins, Milin & Ramscar, 2017).

2. Lifelong learning through constant exposure the linguistic distribution guarantees that speakers' individual samples of linguistic codes must inevitably vary enormously (Keuleers et al, 2015; Ramscar, Sun, Hendrix & Baayen, 2017). This seems to indicate that individual probabilistic models will diverge considerably, such that the various models learned by a language's individual speakers will tend to be anything but *systematic*. It is important to be clear on this point: In the communication systems defined by Shannon, the provision of shared codes ensures that there is <u>no difference</u> between the average probability model of the system and the individual probability models of any two sender / receivers A and B. If we extend this idea to human communication, then where the difference between the average probability model of the system and the probability models of any two speaker / hearers is low, the systematicity of the communicative code will be high. As the difference between the average probability model and the probability models of any two speaker / hearers increases, the systematicity of the code will decrease; and at some point, these differences will tend to falsify talk about 'languages as probabilistic systems.' Since what is currently known about the way human communicative codes are distributed, sampled and learned strongly appears to indicate that the systematicity of human communicative codes will be very low, this in turn begs the question of what it actually means to call natural languages probabilistic systems.

---

[2] In communications engineering, the process by which messages are mapped to source symbols is called *source coding*. A source code is usually configured so as to optimize use of transmission bandwidth, and is a vital part of any communication system. *Channel coding* aims to increase the reliability of data transmission in the presence of system noise by adding *redundancy* to a coded vector of source symbols (at the cost of a reduction in information rate). Shannon's (1948) demonstration that *source* and channel *coding* can be addressed independently in the design of communication systems was an important engineering contribution. However, although the linguistics literature abounds with talk of channel noise and redundancy, it is not at all clear how to make sense of the idea of channel codes in relation to human communication. For example, while redundancy in communication systems can be readily defined in relation to the properties of the source code, since the properties of human communicative codes are yet to be fully understood, any talk of redundancy in relation to natural language is somewhat redundant in itself. Accordingly, in what follows, I will assume that since natural languages can be characterized as codes that allow messages (that relate to discriminable human experiences) to be mapped onto discriminable speech contrasts (source symbols), they are best analogized to source codes.

3. While the distribution of word lengths and frequencies etc. observed in natural languages is as Zipf (1935; 1949) noted, skewed, the power-law distributions that appear to characterize language at various levels of description are not those one would predict from the vast literature on communication and coding theory, simply because power laws are <u>not</u> the most efficient distribution for codewords in variable length codes (see e.g., Gallager & Van Voorhis, 1975).

4. Both Hartley (1928) and Shannon (1948, 1956) note that communication theory is guided by deductive principles. At heart, information theory assumes a discriminative process of uncertainty reduction, in which communication is a sequential process that leads to the elimination of a receiver's uncertainty about the identity of a sender's message, based around a code that is designed to maximize the discriminability of one message from another while minimizing the cost of signaling. The eliminative / deductive nature of this process (Shannon, 1956), which presupposes that the code is *shared*, seems to be opposed to, rather than compatible with, the longstanding linguistic principle of compositionality, which holds that meanings are composed (inductively) from their *independent* parts.

In other words: while information theoretic models of communication (and their operations) rely fundamentally on the existence of <u>shared codes</u>, it seems that human communicative codes are not shared in anything like the same way; While information theoretic models of communication describe a number of constraints on optimal codes, it seems clear that, notwithstanding the many claims that made to the contrary in the literature, these constraints are not obviously met by linguistic codes; And finally, while information theoretic models of communication suppose a deductive process (Shannon, 1956), linguistic theories, even when they claim inspiration from information theory, tend to make 'inductive' assumptions about the way that meanings are communicated.

While these problems differ in the degree to which they are critical to the endeavor of applying information theory to human communication, one thing follows clearly from them: unless it can be explained how the users of natural languages converge on a model of linguistic probabilities that is shared in a way that is at least *sufficient* to satisfy the constraints imposed by information theory (that is, unless it can be explained how natural languages specify the model that defines *information* in human communication), the analogies drawn between language use and information theory, or talk about languages as probabilistic codes, will be largely meaningless. Indeed, unless these challenges can be met, information theoretic approaches to human communication are unlikely to amount to anything more than another theoretical bubble.

In what follows, I will review these problems in more detail and then show how they can be addressed and largely resolved in a domain that is typically ignored by linguists, yet has proven to be a huge stumbling block for theoretical accounts of how natural languages facilitate meaningful communication: systems of personal names (indeed, this domain also seems particularly problematic for information theoretic approaches to human communication). The data and analyses presented below show how the distributional structure of the personal name systems of the world's languages conforms remarkably closely to the predictions of information theory, and indeed, that it appears that the structures of personal name systems have socially evolved so as to be remarkably well adapted to meeting the communicative challenge posed by the incompleteness of human communicative codes. I will also show how the communicative function of names makes clear that while names typically have combinatoric forms, the semantic role played by names in comprehension is discriminative: Names serve to reduce uncertainty about the identity being communicated, they do not 'convey' a name's 'meaning.' Finally, I will briefly describe a range of findings showing that the structures that support a discriminative process of communication are clearly evident in other linguistic domains, such as systems of English nouns and verbs, and the structure of spoken English.

### 1.2. Informativity, coding and the nature of linguistic distributions

An important result of information theory is that variable-length codes (encoding source symbols using a variable number of bits) can allow a lossless source code to be compressed arbitrarily close to its entropy, enabling the efficiency of communication to be maximized close to its theoretical limit (see e.g., Shannon, 1949; Huffman, 1952; Golomb, 1966; Rissanen, 1984). The benefits of efficiency that these codes bring can be intuitively grasped by means of a linguistic example: if the English number *one* is easy to articulate, whereas the articulation of *five* and

*twenty* require successively more effort, then if we skew the distribution of numbers, so that *one* is highly frequent, and *five* and *twenty* successively less frequent, we can help to minimize the average effort speakers will expend articulating English numbers. Moreover, as the distribution and articulation of the numbers *one*, *five* and *twenty* suggest, natural languages do appear – at least superficially – to be organized in a way that balances the requirements of discriminating lexical contrasts against the effort involved in their articulation (Lindblom, 1990).

For example, studies have repeatedly shown how the frequency of words is reliably associated with their length. Less informative words (which are more frequent, e.g., *one*) are reliably shorter than more informative words (which tend to be less frequent, e.g., *nineteen*; see e.g., Zipf, 1935; 1949; Piantadosi, Tily, & Gibson, 2011). Studies of speech production further support the suggestion that there is a relationship between informativity and resource allocation in the structure of natural languages. Words that are more predictable (and thus less informative) in context are more likely to undergo articulatory 'reduction' (i.e., the 'vowel quality' of words decreases as they become less informative, while syllables are articulated with shorter durations, and become more likely to be 'deleted' etc., for reviews see Bell, Brenier, Gregory, Girand & Jurafsky, 2009; Seyfarth, 2014),[3] while disfluencies and other difficulties in language processing are strongly associated with increases in lexical information (Howes, 1957; van Rooij & Plomp, 1991; Westbury et al, 2016).

The frequency distributions of words in natural languages (including numbers like *one, five* and *twenty*; Benford, 1938) further resemble variable length codes in that they are skewed. The distribution of type / token frequencies in any large linguistic sample follows Zipf's law, so that around half of the tokens will represent a 100 or so high-frequency types ('*and*', '*the*'), while the other half will comprise large numbers – perhaps millions – of low-frequency types ('*comprise*', '*corpus*'), some half of which will occur only once in the sample (Baayen, 2001), resulting in a long-tailed power law distribution of word types by their token-frequencies. This fact has been further taken to support the idea that the organization of linguistic codes has evolved – in response to socio-communicative selection pressures – to optimize communication (Zipf, 1935; 1949), and numerous theories have sought to explain how power distributions assist in the optimization of linguistic communication (see e.g. Mandelbrot, 1966; Manin, 2009).

However, there is a clear and critical difference between the distributions observed in natural languages and those specified in optimal communicative codes. Numerous results in coding and information theory have shown that members of the exponential family of distributions (including the discrete counterpart of the exponential distribution, the geometric, Gallager & Van Voorhis, 1975) better serve to maximize the communicative efficiency of codes than power law distributions. Since these findings make clear that an optimal communicative distribution of discrete elements such as words ought to be geometrically distributed, it follows that Zipfean lexical distributions are far from optimal (a point made clear by a direct comparison of the coding efficiency of code-words using a geometric as compared to a power-law distribution).

---

[3] It seems worth noting that the standard ways of describing these phenomena in linguistics sit somewhat awkwardly with the statistical reality of natural languages, since they suggest that that the majority of forms people vocalize are 'reduced' from a notional canonical form, and that it is the articulations of the minority of forms that are rarely encountered that are somehow 'normal.'

Further, it has also been noted that empirical fits between power laws and linguistic distributions are often poor (Baayen, 2001), such that linguistic samples are often better described by fits to other distributions (Clauset, Shalizi & Newman, 2009), and it has been suggested that linguistic power law distributions may in fact represent mixtures of other distributions (Farmer & Geanakoplos, 2006; Gerlach & Altmann, 2013). These observations raise an obvious question: do power laws best characterize the functional distributions of forms in natural languages, or, since coding theory predicts that functional distributions of codewords ought to be geometric, does Zipf's law describe a mixture of actual functional distributions?

### *1.3 Coding and the system solution to communication*

Shannon's (1948) solution to the problem of communication over a noisy channel[4] accepts channel noise as a given and focuses instead on the use of communication systems to overcome channel limitations by detecting and correcting the errors it introduces (MacKay, 2003). The system envisaged by Shannon comprises an information source and a destination, and adds an encoder before the communication channel and a decoder after it. The encoder encodes source messages using the channel code (adding redundancy to the original message). The channel then adds noise to the transmitted message, yielding a received message that comprises a mixture of the source message and noise. The decoder then uses the known redundancy introduced by channel code to discriminate the original source message from both the added noise and other possible source messages.

In applying communication theory to language, it is important to be clear about what Shannon's system solution does and does not involve:

1. The system solution requires a source encoder and a decoder that both have access to the source and channel codes, which define the scope of the possible messages that can be transmitted.

---

[4] See McKay (2003) for a detailed and accessible expansion of all these points.

2. The system solution is not at all concerned with the meaning of messages. The goal of Shannon's system solution is that the receiver be able to successfully reconstruct the source message from the received message by discriminating the source message from other possible messages that might have been selected, and from noise introduced by the communication channel.
3. The purpose of the decoder is not to interpret or expand on the source message in any way. It is simply to reproduce the message at the destination with no loss of signal content.

The difference between communication as envisaged in information theory and the compositional view of language traditionally envisaged by linguists ought to be very apparent. Shannon's system solution is designed to allow the receiver to guess the message encoded at the source with a high degree of accuracy. The use of the word 'decode' in information theory does not correspond to the idea of a listener decoding a speaker's meaning in the standard model of language; it merely corresponds to the listener being able to use a common code to successfully discriminate the message a speaker did send from those messages a speaker might have sent.

This last point highlights another important difference between the kind of communication system envisaged by Shannon and what we actually know about natural languages. Shannon's solution to the problem of communication over a noisy channel assumes that every part of the system has access to shared codes. By contrast, natural languages are learned, and learning any given aspect of a linguistic code appears to rely on its being attested to in the learner's experience. That is, it seems reasonable to assume that a child can only learn the English word 'dog' when its place in the code is attested to by someone saying, '*dog*' (Ramscar, Dye & Klein, 2013). However, when combined with the observed distributions of words described above, this presents a problem. Because most word types occur very infrequently in the code, it follows that at <u>any</u> stage in the life of any learner learning a code by exposure to it, many of the word types present in the code will not have been attested to. Moreover, studies of language sampling (e.g., Ramscar et al, 2014; Keuleers et al, 20165) and morphologically rich languages (e.g., Blevins, 2016; Blevins, Ackerman, Malouf & Ramscar, 2016; Blevins, Milin & Ramscar, 2017) clearly show how the distribution and productivity of linguistic codes both guarantee that previously unattested forms will be encountered frequently in everyday linguistic communication.

*1.4 Codes and meanings*

Another difference between information theory and traditional linguistics that communicative approaches to language tend to elide relates to the relationship between linguistics signals and meaning. Contemporary approaches to language – whether generative, cognitive or communicative – all tend to adhere to the principle of compositionality, in which the meanings of strings are inductively composed from their parts. By contrast, both Hartley (1928) and Shannon (1948; 1956) – the founding fathers of information theory – took pains to point out point out that information theory describes a discriminative (deductive) process. Information theory comprises a series of methods for optimizing the process of communication based on uncertainty reduction, in which signaling is a sequential process that serves to gradually eliminate a receiver's uncertainty about the identity of a sender's message, based on codes that are designed to maximize the discriminability of one message from another while minimizing the cost of signaling.

From this perspective, insofar as the process of communication described by information theory can be related to natural language, the words in an utterance are not be considered to be 'carrying' or 'encoding' meanings (since these ideas ultimately violate coding theory, Ramscar, Yarlett, Dye, Denny, & Thorpe, 2010). Instead the semantic function of words is to reduce a hearer's uncertainty about what a speaker intends to mean through a process of elimination (Ramscar & Port, 2016). That is, information theory adopts a very different perspective towards the relationship between signaling and meaning than the compositional accounts typically embraced by linguists. This perspective has much to recommend it, not least because as Ramscar & Port (2015) point out, it is far easier to provide a computational account of how a learner might acquire the ability to discriminate between, say, *foxes* and *dogs,* or *two* and *three,* than it is to explain how someone might acquire a representation of an atomic meaning of *fox* or *three* (or even say clearly what the latter is supposed to be, Wittgenstein, 1953). Perhaps accordingly, the contemporary cognitive models that best fit and predict human performance on tests of categorization (e.g., Nosofsky 1991, 1992; Love, Medin & Gureckis, 2004) do so by not making use of discrete concepts at all. Instead these models all focus on discriminating appropriate category labels – or matches between labels and exemplars – in context from continuous conceptual spaces (Ramscar & Port, 2015), a process that corresponds to learning the aspects of an experiential space that are best discriminated by a given label (Ramscar et al, 2010).

Moreover, although meaning compositionality is often taken as a given in linguistics, it has long been clear that compositional, (i.e., realist) accounts of word meanings encounter many deep problems when it comes to explaining what discrete word meanings (the basic elements of compositional semantics) are actually supposed to be, or how they are learned (Wittgenstein, 1953; Quine, 1963 Fodor, 1998). For example, although the 'meanings' of words like *two* and *three* seem obvious at first blush (*two* 'means' something the set of exactly two countable objects and *three,* 'means' the set of exactly three objects; see e.g., Carey, Shusterman, Haward & Distefano, 2017), not only are these and other similar 'definitions' tautologies (in that they use the word *three* to 'explain' the meaning of 'three'), but if we look at the way that *two* and *three* are actually used in English – by consulting a large corpus – we find that, along with the words for other highly frequent cardinal integers such as *four, five, six* and *seven*, *two* and *three* are more often than not used to talk about *time* (the countability of which is debatable) in ways that are more often than not far from exact (Ramscar, Matlock & Dye, 2010).

The distributional facts illustrate just one of many known problems that compositional ideas about meaning inevitably incur.

(1)     *Dinner will be ready in two minutes*

If we assume that the speaker in (1) does not intend 'two minutes' to mean exactly 120 seconds, then what exactly does *two* mean here? And, if one is to suppose that this use of 'two' is somehow 'metaphorical' (which also tends to be defined by tautology; Ramscar & Pain, 1996) then how is one to account for the fact that *two*'s metaphorical meaning is used more often than its supposed 'literal' meaning; and how is one to give an non-tautologous account of *two*'s literal meaning in this context, etc.? [5]

Notably in this regard, some of the biggest problems posed by the idea of meaning compositionality have been encountered in relation to personal names (Frege, 1892; Russell, 1919; Searle, 1971; Donnellan, 1972; Burge, 1974; McDowell, 1977; Boersema, 2000). Although the problems posed by *two* in (1) above are widely acknowledged, there is also a broad consensus in contemporary linguistics and linguistic philosophy that these problems are – somehow – solvable in compositional terms as a result of the existence of underlying concepts (whatever 'underlying concepts' may actually be; see Ramscar & Port, 2015).

(2)     *John will be here in two minutes*

Yet, when it comes to explaining what the compositional meaning of *John* in (2) is supposed to be, let alone how this meaning is acquired and represented psychologically, then although it is generally agreed that names such as *John* pose particular problems for compositional accounts (albeit that, as Wittgenstein, 1953, points out, on close analysis the problems posed by *two* and *John* are ultimately the same), there is little consensus as to how these problems are to be solved in even the broadest theoretical terms (see Gray, 2014; Fara, 2015, for reviews).

By contrast, providing an account of the semantic function of names such as *John* in communicative terms turns out to be something of a straightforward matter. However, since this explanation relies in turn on detailed accounts of how natural languages satisfy the various distributional and systematic requirements of information theory, it will first be necessary to provide these. Accordingly, Section 2 of this paper briefly reviews a proposal by Blevins, Milin and Ramscar (2017), who suggest that the productive aspects of morphological systems can be seen as serving to offset the problems that arise out of the incompleteness of natural language codes. Section 3 then addresses how a communicative perspective can be brought to bear on systems of personal names, an aspect of natural language this usually thought of as being both unstructured and hardly communicative at all. It will show how the name systems of the World's languages share a highly structured, historically universal form that appears to have evolved to satisfy the communicative requirements described above. It will also describe how the apparent differences in the structure of names systems across languages today can be accounted for in straightforward terms in the light of the information structure of these systems, and the way that different forms of name legislation have interacted with them (linguistically, personal names are somewhat unique in that their form tends to determined by legal as well as communicative pressures; Scott, 1998). Section 4 will then make good on my promise of providing a detailed account of the how names serve their semantic function in communication. The remaining sections will then describe how this account can be extended to other aspects of human communication.

---

[5] The intractability of these problems has prompted some advocates of compositionality to claim that the meanings of all words, even *bureaucrat* and *carburetor*, are innate (Fodor, 1998; Chomsky, 2000, see below).

## 2. Partial attestation and the incompleteness of human communicative codes

While it is clear that no speaker has full access to the whole communicative code of any natural language, when taken together with the fact that human communicators learn, the morphologically productive nature of many aspects of natural languages offers an example of how human communicative codes might approximate Shannon's system solution in a way that is <u>sufficient</u> to serve users' communicative needs. As Blevins, Milin & Ramscar (2017) point out, although morphological productivity might seem to increase the probability that unattested forms will occur in a system at first blush, in practice the members of productive morphological paradigms whose forms are most likely to be unattested (because they are infrequent) can inevitably be inferred from mutually informative neighbors. That is, because natural language users are learners, if partial information is appropriately structured it may be sufficient to allow for successful communication. Accordingly, the existence of productive regularities in morphological systems can be seen as a functional solution to many communicative problems that might otherwise impact any speech community whose members possess only variable and incomplete samples of a code, because in both comprehension and production the existence of

'regular' form classes enables unattested forms to be *inferred* from partial samples (Blevins et al, 2017; Ackerman & Malouf, 2013; Ramscar et al, 2013a).

From this perspective the coexistence of regular and irregular patterns can be understood in terms of a trade-off between two opposing pressures on communication: *predictability* and *discriminability*. Irregular forms are both frequent and well discriminated, and thus they serve to emphasize important communicative contrasts that will be less saliently marked in regular patterns. Additionally, as well as making the communication of high-frequency signals more efficient, irregular forms of may also serve to enhance the learnability of the system by maximizing the discriminability of important contrasts (Ramscar et al, 2013a; 2010). By contrast, while regularity necessarily entails less discriminability between forms (reducing communicative efficiency), the existence of regular lexico-morphological neighborhoods serves to compensate for the fact that an individual will always have only partial access to a communicative code: While the forms of low frequency items will often be unattested in experience, the fact that the inflectional patterns they follow can be deduced their form neighborhoods means that they will be *partially attested,* in that their forms will be predicable within the system (Blevins et al, 2017).

Convergent support for the suggestion that regular and irregular patterns reflect a *predictability* and *discriminability* trade-off within linguistic codes comes from studies of grammatical gender (Dye, Milin, Futrell & Ramscar, 2017). Grammatical gender – i.e., the linguistic practice of assigning nouns to discrete classes that share some morphological properties – has traditionally been derided as being both unsystematic and functionless by linguists (Bloomfield, 1933; Kilarski, 2007; Corbett, 1991), however for a species as disposed towards technology as *homo sapiens*, nouns pose problems that dwarf those of many other parts of speech (Dye, Milin, Futrell & Ramscar, 2018), namely that humankind has a remarkably propensity for inventing new things, and these new things usually need names. As a result, the set of named objects in most languages is vast and growing all the time (such that any given speaker will never encounter many of the nouns in their native language). In an analysis of the German gender (noun class) system, Dye et al (2017) found that the gender of lower frequency German nouns is predictable from the lexical-semantic contexts in which they occur (see also Schwichtenberg & Schiller, 2004). Further, other parts of speech that appear in noun contexts (e.g. adjectives) are often marked for gender in German, such that this system serves to increase the overall predictability of the form (and indeed, the content) of noun phrases in context. By contrast, it is the way that gender distributed across high frequency German nouns that makes the system appeared 'arbitrary.' This is because gender serves a discriminative function for these nouns: as the likelihood that two high frequency German nouns will occur in the same lexical contexts increases, so does the likelihood that their gender will *differ*, increasing the discriminability of these nouns in context (Dye et al 2018, show that in English, a largely non-gendered language, the distribution of pre-nominal adjectives serves a similar function).

In other words, natural languages differ from digital information systems in that their codes are adaptive systems that change constantly as individuals and communities develop and learn (Ramscar et al., 2014; Ramscar & Port, 2016). However, although this means that it is impossible for any given speaker to have full access to the source code of a natural language (or at least, any modern code), it follows that if information theoretic approaches to language are on the right track, we should expect linguistic codes to be structured along the lines just described at every level of description. That is, we should expect that the structure of forms known to all speakers should tend to maximize discriminability in communication, while the distributional structure of less frequent forms should support the productive regularities that appear to support the plasticity of natural languages while at the same time maintaining their integrity as codes.

It is thus worth making explicit a point clearly implied in the foregoing: the approach to the linguistic environment that communicative theories ought to take is the opposite of that put forward in 'poverty of the stimulus' arguments put forward by other linguistic theories. Many linguists have abandoned falsifiability (and thus science) in their theories by taking the position that evidence indicating that a theoretical postulate is unlearnable does not falsify the postulate, but rather supports the claim that it must (somehow) be innately specified in the human genome. By contrast, communicative theories must, by necessity, posit that there is a great deal of statistical structure in the linguistic environment. It follows from this that both the discovery of *and* the failure to discover postulated structures can allow predictions from communicative theories to be supported, modified or rejected using the standard methods of empirical scientific enquiry. With this in mind, I turn to the communicative structure of personal names. Children are usually given names that are deliberately chosen. Yet despite the old adage to, 'choose your words carefully,' the latencies associated with spontaneous speech suggest that most communicative behavior does not involve the kind of deliberative thinking associated with active choice (see Lee, Seo & Jung, 2012, for a review). Given this, and given traditional linguistic assumptions about names being barely communicative at all, one might reasonably ask whether name systems actually have a communicative structure at all. As I shall now show, the name systems of the World's languages have a surprisingly amount of structure – and a surprising amount of structure in common.

# 3. Name systems in human communication

## 3.1 Incompleteness and the information cost of naming

Most words in natural languages – e.g., *two, red, walk, dog, idea*, etc. – are semantically generic, in that they can be applied to a wide range of individual objects and events. However, the semantic function of personal name is, at least in part, *sui generis*, in that one of the functions of a personal name is that of discriminating an individual from their societal peers. In very small social groups, this function might be realized by employing a unique name token for each individual (see e.g., Hvenekilde, Marak & Burling, 2000). However, unique name token systems will inevitably encounter insurmountable information processing challenges as populations grow. Furthermore, a unique name token system would rapidly run into insurmountable problems with regards code sharing: while the number of words for *run, drink, spoon* or *water* in any given communicative code appears to be little influenced by the size of its population of speakers, given that the most obvious function of a personal name is that of discriminating an individual from their peers, it follows that the number unique names that a code must serve to discriminate must, to some extent, inevitably scale with its population of speakers. This means in turn that, given that all human communicative codes employ a very limited set of acoustic contrasts, as the sizes of community of speakers grows, individual names in a unique token system would soon become insurmountably long, or insurmountably confusable; and in either case, their unpredictability would likely pose insurmountable problems to communication.[6]

Perhaps accordingly, all of the world's major languages appear to have evolved similar structural solutions to the communicative challenges raised by the functional requirements of naming. Rather than employing unique tokens, identifiers for individuals – names – are formed from sequences of hierarchically structured name tokens that are rarely unique themselves (Ramscar et al, 2013e). By deploying sequences of name tokens, the name grammars of the world's languages allow for large sets of more or less unique identifiers to be constructed out of far smaller sets of name words. These systems thus allow individuals to be identified by relatively unique name sequences while avoiding the scaling problems associated with systems of unique name tokens. Indeed, given that name grammars serve to reduce semantic uncertainty (about an the identity of the individual being communicated) across a vast range of contexts by combining far smaller sets of form contrasts into unique permutations, functionally they might be seen as a microcosm for natural languages in their entirety.

---

[6] Indeed, it would seem that this point applies to any aspect of a communicative code.

Although a full analysis of the historical evolution of name grammars is outside the scope of this analysis (see Ramscar et al, 2013e for a sketch), two important aspects of their development are important to clarify here: first, from a historical perspective, the patronymic 'family names' observed in modern name grammars are remarkably recent invention; and second, from an information theoretic perspective, the order in which name tokens appear in signals is a <u>critical</u> determinant of their communicative function.

*3.2 Patronyms from a historical and cultural perspective*

Humans have been communicating linguistically for some 100,000 years or so. However, for current purposes it is important to note that for the vast majority of this time, the overwhelming majority of people had just a single 'formal' name taken from a small set of name tokens (Smith-Bannister, 1997). This name could then further modified to satisfy a speaker's communicative goals through the addition of descriptors, nick names etc. In practice, this tended to mean that given name tokens served as the first part of a more discriminative name sequence (e.g., the sequence *Aristotle Stagiritis* – 'Aristotle from Stagira' – serves to discriminative a philosopher with the given name *Aristotle* from other people also given the name *Aristotle*).

Some two thousand years ago in what is now China, name sequences had come to formalize heredity (of either blood or property) among the landed aristocracy, however, these more complex naming conventions applied to only a small, albeit powerful, minority of the population (Du, Yuan, Hwang, Mountain & Cavalli-Sforza, 1992). Then around 200 BC, the Qin dynasty – the world's first bureaucracy – enacted laws that appear to have turned a more flexible 'native' system of names (a given name token plus modifiers) into a fixed, regulated system. These naming conventions transformed given names into patronyms that, in particular, determined the names given to children born after a male named X married a female named Y (prior to this, female forms of name tokens were marked for gender by an extra character – *shi* – in a manner similar to the way that Western first names often have make and female forms, e.g., *Henry / Henrietta, Franz / Franziska*, etc., Du et al, 1992). Under the new regulated system, a female did not change her name on marriage, but all children tool their father's given name as their first name, following the practices of the aristocracy (Du et al, 1992). These regulations led in turn to the less formal systems of descriptors, nicknames etc. being repurposed into formalized 'given names.'

By contrast, in England (and the West more broadly) given name tokens continued to function as given names, and when naming practices began to be formally regulated in the early modern period, it was the modifying tokens that had previously been employed to add discriminatory information for communicative purposes that were repurposed and employed as patronyms (Smith-Bannister, 1997). The semantic consequences of transforming transparent nicknames and descriptors into opaque patronyms is particularly evident where the former descriptor was itself a patronym, as for example with adding 'Johnson' to describe a son of John. In the modern English-speaking world, there are now hundreds of thousands of women whose fathers are named *David, Michael, Dwayne, Steven* etc., and who are definitely not sons, yet nevertheless go by the hereditary name *John<u>son</u>* as a consequence of naming legislation.

Because discussions of names (and translation conventions) largely taken the existence of patroness (or 'family names') for granted (ignoring both their recency and the degree to which they were often arbitrarily imposed on populations Scott, 1998), and because an English name comprises a first given name and a last patronymic family name, 'last name,' 'patronym' and 'family name' are often used interchangeably. Because of this, and despite the fact that a Chinese name comprises a first patronymic family name and a last given name, Chinese patronyms are often (misleadingly) referred to as 'last names' in English.

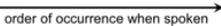

**Figure 1.** Two forms of the same name: that of actor Bruce Lee, aka Lee Jun-fan. In English, Lee's *first name* was Bruce (because in English, the order in which Lee's name is spoken is Bruce followed by Lee), whereas in Mandarin his *first name* was Lee (in speech, Lee precedes Jun-fan).

In other words, when say English and Chinese names are compared, these comparisons usually tend to focus on the social-signaling aspects of names, e.g., whether name tokens are inherited or given. However, this serves to confuse and obscure their information function, which is determined by the <u>sequencing</u> of name tokens in communication, and the degree to which they reduce uncertainty about the identity being communicated. Because the focus of this paper is the information function of names, I will use *first name* to describe the part of a name sequence that usually comes first as a name is spoken, and *last name* to describe the part of a name sequence that usually comes last (Figure 1).

*3.3 The information structure of names in different languages*

Names in modern Chinese (a family of Sino-Tibetan languages, Matisoff, 1991; Handel, 2008) and modern Korean, a language isolate (Song, 2006) typically comprise two or three elements (Kiet, Baek, Kim & Jeong, 2007). For example, as a full Korean name (3) is encountered in speech, these comprise sequentially: first, one of a small number of family names; then second, (usually but not always) a clan or generational name; and then third, a given name. The size of the sets that each name element is drawn from increases as names successive unfold, so that Korean and Chinese names can be seen to serve as hierarchical decision (or discrimination) trees: each element iteratively increases the degree to which a given individual is identified, while the structure reduces the number of alternatives that need to be considered at each step.

| (3) | Family Name | Clan / Generation Name | Given Name |
|---|---|---|---|
|  | Least Uncertain | More Uncertain | Most Uncertain |
| e.g., | *Baek* | *Seung* | *Ki* |

In many Western languages, such as English (a member of the North Germanic family of languages, Faarlund & Emonds, 2016), family names form the last part of a personal name. However the overall information structure of English and Korean names are remarkably similar. English given names (which occur first, sequentially, as a full name is encountered in speech) are drawn from far smaller sets than English family names (in a sample of 6.2 million records from the 1990 US census there were 16 surnames for every given name), such that the given name *'Michael'* is highly frequent (it was common to around 4 million English speaking US males in 1990). By contrast, *'Ramscar'* (the author's family name, derived from a location in Yorkshire, Ramskir, 1973) is far less frequent, such that the overall likelihood of each element in (4) iteratively decreases across the name sequence in the same way as in (3), with each successive element increasing the degree to which the author is uniquely identified.

| (4) | Given Name (s) | Family Name (s) |
|---|---|---|
|  | Less Uncertain | More Uncertain |
| e.g., | *Michael* | *Ramscar* |

*3.4 The distributional structure of Chinese and Korean names*

Historically Sinosphere first names were drawn from a small set comprising around 100 or so commonly used name tokens (Yuan, 2002; the literal translation of colloquial Chinese expression for the common people – *'Bǎijiāxìng'* - is 'the hundred names'). The set of Korean first names was similarly restricted in size (Baek, Kiet & Kim, 2007), and studies have shown that the distributions of these two functionally identical sets of lexical items violate Zipf's law: unlike family names in most of the World's languages (whose distributions do appear to follow power laws) Chinese and Korean family names are geometrically distributed (Yuan, 2002; Kim & Park, 2005; Kiet, Baek, Kim & Jeong, 2007; Guo, Chen, & Wang, 2011; see also Figure 2). [7]

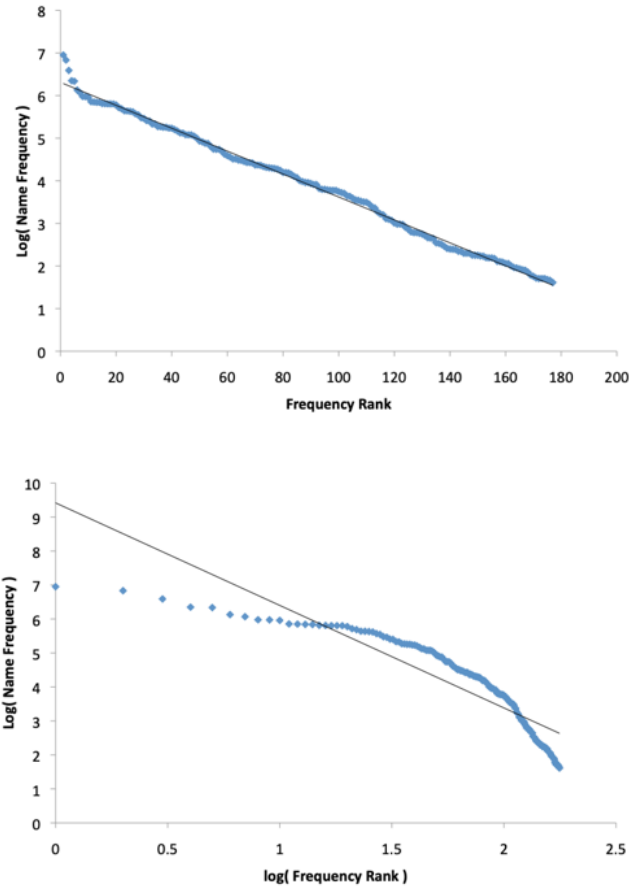

**Figure 2. Top:** Log frequency x frequency rank plot (linear = exponential) of first names (minimum frequency = 40) in the 2000 South Korean Census; $R^2=.99$.[8] **Bottom:** Log frequency x log frequency rank plot (linear = power law) of the same data; $R^2=.81$.

---

[7] Baek, Kiet & Kim (2007) show that if one assumes an original distribution of names, along with parameters for migration, birth and mortality rates, the different shapes of Western and Sinosphere family name distributions can be modeled as arising out of differences in the rates at which new names are introduced into society.

[8] http://www.nso.go.kr/board_notice/BoardAction.do?method=view&board_id=77&seq=10 retrieved from https://web.archive.org/web/20071103060323/ November 16th 2016.

As I noted above, geometric (exponential) distributions enable the average information in a set of codewords to be minimized (such that we ought to expect an optimal communicative code to employ geometric rather than Zipfean distributions). Accordingly, if we assume that the function of a sequence of name tokens is that of reducing uncertainty about an identity in communication, then it appears that although Korean and Chinese are not 'genetically' related languages (Song, 2006), the distribution of first names in both language suggests that their name systems may be optimal for the purpose of communicating about individuals in this way. This idea also raises a number of questions: Are Korean and Chinese first name distributions really unique? Do their distributional forms really reflect hereditary constraints imposed the hereditary factors associated with families (as Baek, Kiet & Kim, 2007 argue) rather than those of communication? And if these distributions have been shaped by communicative constraints, then why is it that the same constraints are not also reflected in the name distributions of other languages?

*3.5 The distributional structure of Vietnamese first names*

Although a great deal of data related to name distributions exists, its study is complicated by a number of related factors: the fact that the name laws that made fixed legally defined names ubiquitous to modern life are intimately linked to the development of the bureaucracies that use these names to make citizens and their activities more 'legible' to the governments of modern nation states (Scott, 1998) means that name data tends to be the preserve of states; meanwhile, the fact that responsible states require publicly released data relating to individuals to be anonymized (see Ohm, 2010 for a review) inevitably conflicts with the fact that the semantically sui generis nature of names (and especially fixed, legally defined names) often makes the identification of individuals from their full names a trivial matter.

As a result, statistical studies of names rely on partial or historical data. For example, in Korea many families have for centuries followed the old Confucian tradition of recording their ancestral trees in special books (called family books). Data from these books has been shown to be invaluable when it comes to estimating the distribution of Korean first names over time (Kiet, Baek, Kim, & Jeong, 2007). In a similar vein, the fact that the overwhelming majority of Vietnamese-Americans sampled in the 2000 census are naturalized citizens born—and named—in Vietnam, makes it possible to recreate a distribution of Vietnamese first (family) from US last (family) name data to see if distributional findings from Korea and China generalize to another unrelated Sinosphere language (Vietnamese is an Austroasiatic language, albeit with significant vocabulary borrowings from Chinese; Sidwell, & Blench, 2011).

To make this estimate, two lists of frequent Vietnamese first (family) names (Lauderdale, & Kestenbaum, 2000; Taylor, Nguyen, Do, Li, & Yasui, 2011) were combined with a Vietnamese names list from Wikipedea. The frequencies for these names in the population reporting as Asian in the US 2000 census were then calculated, and then the 50 most frequent names were extracted. The frequencies of these 50 names and the 50 most frequent first (family) names from the 2000 South Korean census were then normalized and compared.

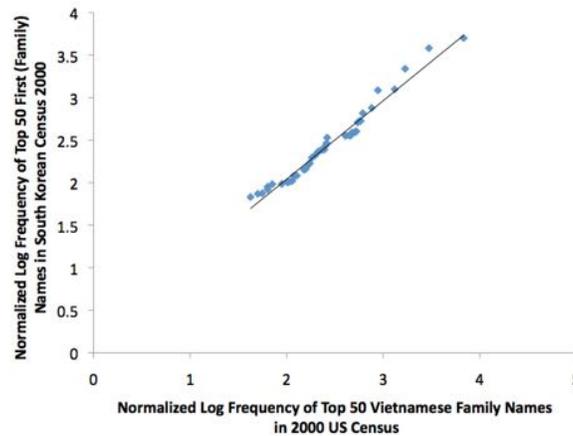

**Figure 3** Point-wise comparison of frequency normalized distributions of 50 most frequent Vietnamese first names in US 2000 census and 50 most frequent first names in Korea 2000 census ($R^2=.97$).

As Figure 3 shows, the distribution of the 50 most frequent Vietnamese first names in the US 2000 census is largely identical to the distribution of the 50 most frequent first names in the Korean 2000 census. Moreover, is further worth noting that whereas the shape of Zipfean distributions changes as a function of sample size, (Baayen, 2001), the shape of these two distributions is the same despite the fact that they represent samples of 38 million South Korean names, versus only 1 million US-Vietnamese names.

*3.6 The distribution of English first names*

Western family names have long been known to be Zipf distributed (Fox & Lasker, 1983; Reed & Hughes, 2003; Baek et al, 2007). However, as noted above, from an information theoretic perspective, the differences in the sequential organization of Western and Sinosphere names suggest that the appropriate communicative comparison ought to be between Western first (given) names and Sinosphere first (family) names rather than Western and Sinosphere family names. However, as Figure 4 shows, analysis of a set of first name data from the 1990 U.S. census clearly indicates that English first names are Zipf distributed.

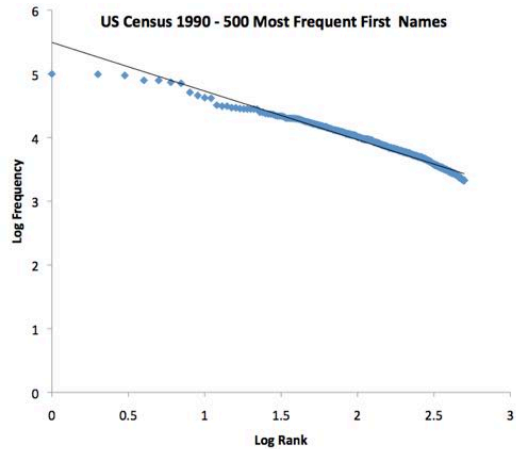

**Figure 4** Log frequency x Log frequency rank plots (linear = power law, $R^2$=.97) of the 500 most frequent English first names in the 1990 US census (this data was extracted from a subset of 6 million records, and released independently of last name data, to conform to anonymity requirements[9]).

However, there are a number of reasons for supposing that the sample plotted in Figure 4 is not an empirically valid communicative distribution: first, because when exponential or geometric distributions are mixed, the result is often a power law distribution (Farmer & Geanakoplos, 2006); and second, because the USA has a population of over 325 Million people. In the unlikely case that someone was able to spend 12 hours a day meeting a new person every 10 seconds, it would take them over 200 years (i.e., considerably more than the human lifespan) to meet every member of the population and hear their name once. Given that no one can actually sample the entire US name stock, it follows that the total stock cannot serve as the basis of any individual's communicative model, or be the cause of any individual communicative behavior. Rather, it seems far more likely that smaller samples serve to influence individual naming decisions, such that the distribution of names in Florida serves to influence individual naming decisions made in Florida more than those made in Oregon, and vice versa; And it follows accordingly that combining the name distributions of, say, Florida and Oregon will paint an inaccurate picture of the distributions that serve to influence actual naming decisions.

---

[9] See https://www2.census.gov/topics/genealogy/1990surnames/nam_meth.txt

However, before considering the implications these factors have for our understanding of modern name distributions, it will be helpful form a picture of the historical distribution of English names *before* their distributions were affected by the imposition of legal restrictions on names, and before the population of English speakers grew to the extent that the aggregate of English names can hardly be called a system. Not least because not only is it the case that name legislation and fixed legal names are remarkably recent developments in the West (Scott, 1998), but also because it seems clear that the distribution of English first names was *remarkably* stable in the centuries prior to the imposition of name laws (Lieberson & Lynn, 2003). It thus seems that a historical perspective may be essential to understanding the full influence these laws have had on Western first name distributions, especially as populations have grown since their introduction.

### *3.6.1 Population effects on the distributions of first names in England*

In the UK, naming laws were only fully standardized in the early 19$^{th}$ Century, following the Births and Deaths Registration Act of 1836, such that the pattern of surnames we know today wasn't fully established until the 20$^{th}$ Century (Hanks, Coates, & McClure, 2011). Although the total population of England fluctuated significantly as a result of plague and famine in the 500 years leading up to this, its level in 1800 was only around 50% greater than it had been in 1300 (Wrigley & Schofield, 1989; Jeffries & Fulton, 2005). Moreover, while this inevitably meant that information in the name system increased (because the number of identities that it encoded increased), it is notable that the distribution of English first names appears to have remained remarkably constant across this period: in every 50-year period between 1550 and 1799, 50% of the boys born in England were given the names *William*, *John* and *Thomas*, while 50% of girls were named *Elizabeth*, *Mary* or *Anne* (Lieberson & Lynn, 2003). That is, 50% of the people of England shared just 6 first names.

However, in contrast to Sinosphere first name distributions, which have remained largely stable over the past several hundred years (Kiet et al, 2007), the distributions of Western first names have changed dramatically after the beginning of the 19$^{th}$ Century. One obvious cause of this change is the inexorable growth in the population of England in this period. Figure 4 plots the distribution of the three most common English male and female first names against the increasing population of England decade by decade across the 19th Century. It reveals a near-perfect relationship between increasing population and declining popularity for the three most frequent male ($R^2$=-.98) and female ($R^2$=-.997) names in this time.

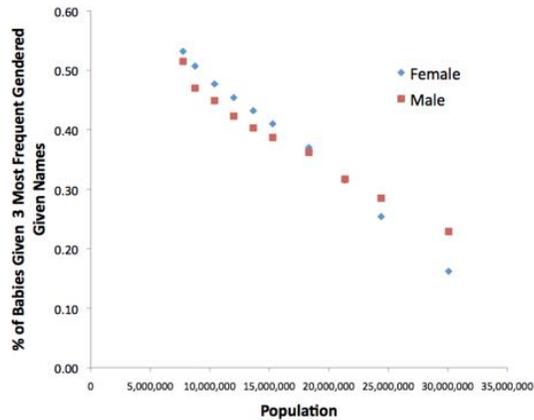

**Figure 4.** The frequency of the top three male and female names were in England 1800 – 1900 (Galbi, 2002) plotted against population size (Wrigley & Schofield, 1989; Jeffries & Fulton, 2005; 1890 data unavailable).

These remarkably systematic changes – which suggest that the distribution of first names in 19th Century Britain was very sensitive to the number of individuals that overall the name system was required to discriminate – raise a further question: how exactly were British names distributed prior to the modern period, and the trend towards continuous population growth?

### *3.6.2 First name distributions in Scotland, 1701-1800*

To make first estimate of the distribution of first names in English speaking communities prior to 1801 a detailed dataset of the names recorded in the registers of four Scottish parishes between 1700 and 1800 was examined (Crook, 2012). Analysis of the data (comprising 23935 records, divided by gender and parish of birth) revealed that in each parish, around 50% of children had one of 3 gender appropriate first names, and around 80% of children one of 10 gender appropriate first names (the names in these sets varied across parishes).

The name stock (the size of the set of first names) in each parish was around 100 (note that the size of the historical name stocks of China and Korea was also around 100, Yuan, 2002; Kiet et al, 2007). Moreover, and further consistent with the first name distributions observed in China and Korea, a comparison of the aggregate distribution of the most frequent 100 male and female first names in the 4 parishes (Figure 3) to an idealized distribution based on the empirical properties of these observations (100 exponentially distributed tokens, with 6 tokens making up 50% of the probability mass) revealed a close fit between the idealized and observed distributions ($R^2=.99$), while further analyses revealed that male and female names were also geometrically distributed when examined individually by gender.

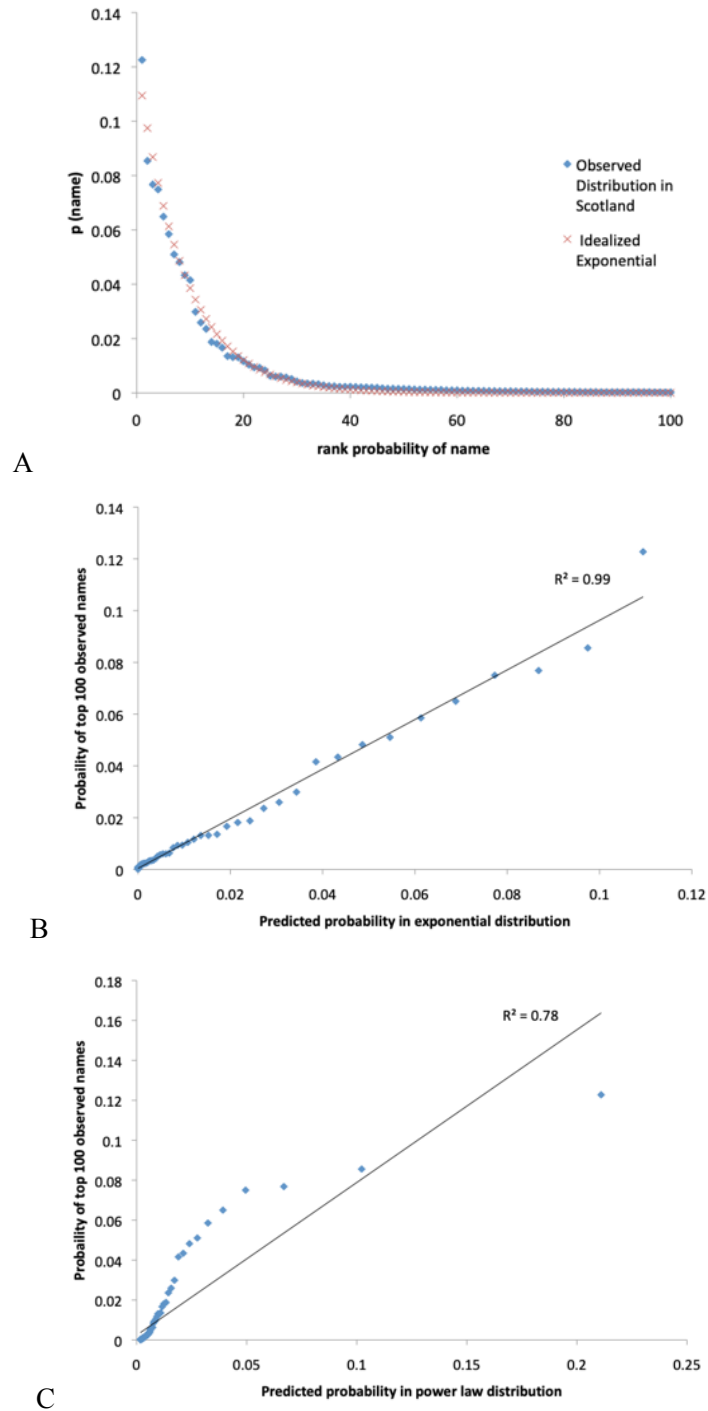

**Figure 5 (Panel A)**: The 100 most frequent given names (98% of the population) 1701-1800 in the Parishes of Beith, Govan, Earlstone and Dingwall in Scotland (Crook, 2012) plotted against an idealized exponential distribution. **(Panel B)**: Pointwise comparison of observed distribution and an idealized exponential distribution based on its empirical properties ($R^2$=.99). **(Panel C)**: Pointwise comparison of observed distribution and an idealized powr law distribution based on its empirical properties ($R^2$=.78).

### 3.6.3 First name distributions in Durham and Northumberland 1701-1800

These results were then replicated on an English sample from the same period, extracted from historical birth records for Durham and Northumberland in England[10] (the sample comprised 25,933 records extracted from a larger data set covering the period 1530-1830). As Figure 6 shows, the name distributions in these two different samples taken from Scotland and England in the 19$^{th}$ Century are largely identical ($R^2$=.98).

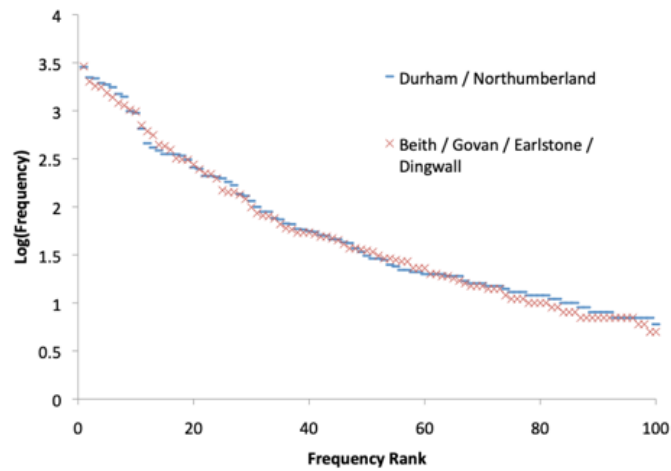

**Figure 6:** Log frequency x frequency rank plots (linear = exponential) comparison of the birth records from Durham and Northumberland 1701-1800 with those of the aggregate records from the 4 Scottish parishes.

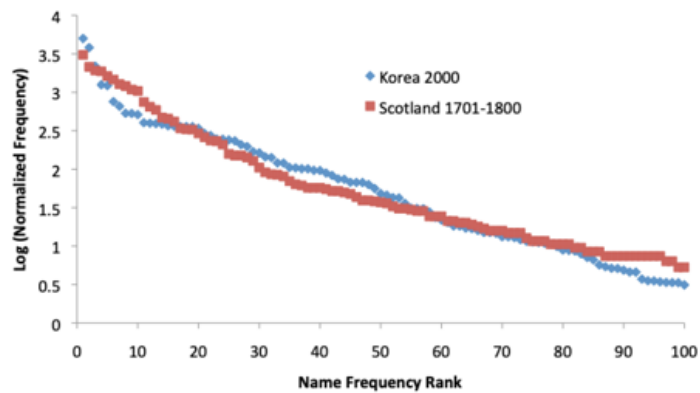

**Figure 7:** Log normalized frequency x normalized frequency rank plot of most frequent 100 first names in South Korea Census 2000 and Scotland / Durham Northumberland 1701-1800 (linear = exponential).

---

[10] Data source http://www.galbithink.org/names/ncumb.txt

Finally, and strikingly, Figure 7 plots a frequency normalized comparison of the distribution of Korean first and Scottish first names, revealing that their shape ($R^2$=.96) and information entropy (Korean=4.7 bits; Scottish=4.8 bits) to be remarkably similar. Given that the communicative function of Korean and Scottish / English first names are exactly the same, but that only the allocation of the former is determined by inheritance, whereas the allocation of the latter is a matter of parental choice, it seems the geometric shape common to these first name distributions is likely to have arisen in response to the constraints imposed by communication (rather than specific factors governed by inheritance, as hypothesized by Kiet et al, 2007)

*3.6.4 First name distributions in modern America*

Studies suggest that historical distributions of first names across Europe were highly similar to those seen in the Scottish and English data analyzed here (Lieberson & Lynn, 2003), a fact that appears to further support the ideas that the collective distribution of first names served a communicative purpose.[11] With this in mind, I return to the question raised above: given that English names are still used communicatively, why are they now Zipf distributed (Figure 4)? As I noted earlier, aggregating names in a country the size of the USA is highly unlikely to result in an empirically valid communicative distribution, and given that mixing exponential or geometric distributions often results in a power law distribution (Farmer & Geanakoplos, 2006), it appears that the name distribution of a smaller English speaking community is likely to provide a better insight into the information structures that serve to influence communication and individual naming choices.

The State of Delaware has a fairly restricted surface area (49th out of 50), and a small (45th) but densely distributed (6th) population. Figure 7 plots the decade-by-decade frequency distributions of first names in the 1910-2010 social security records[12] for Delaware, while Figure 8 plots the *cumulative* distribution of names across this period (i.e., the system that results as each successive generation is added – the process of death was ignored in this analysis because a linguistic identity is not removed from the name communication system each time a person dies, e.g., the identities of *Winston Churchill* and *John F Kennedy* are still frequently signaled many decades after their deaths).

---

[11] Indeed, it may be that the striking similarities in the historical distribution of first names across the World's languages offer an insight into the capacity limits of human communication.

[12] Data from https://catalog.data.gov/dataset/baby-names-from-social-security-card-applications-data-by-state-and-district-of-

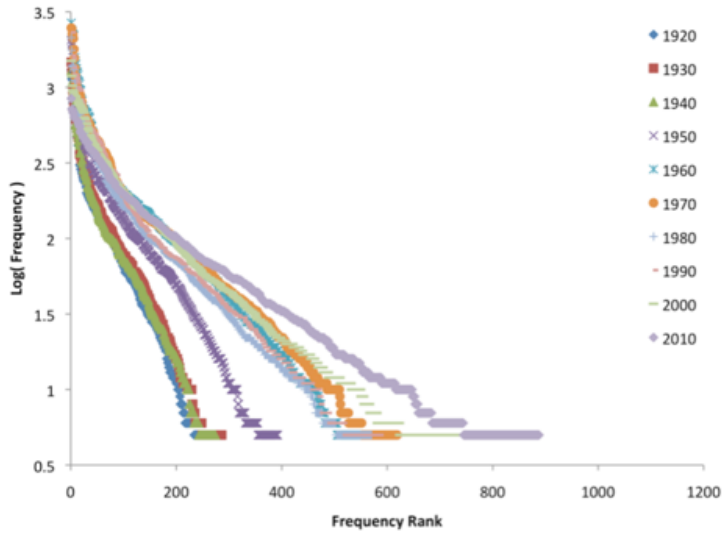

**Figure 7:** Log frequency x frequency rank plot (exponential is linear; all $R^2 > .97$) of the names recorded in the social security records (minimum frequency = 5) in Delaware in each decade 1910-2010.

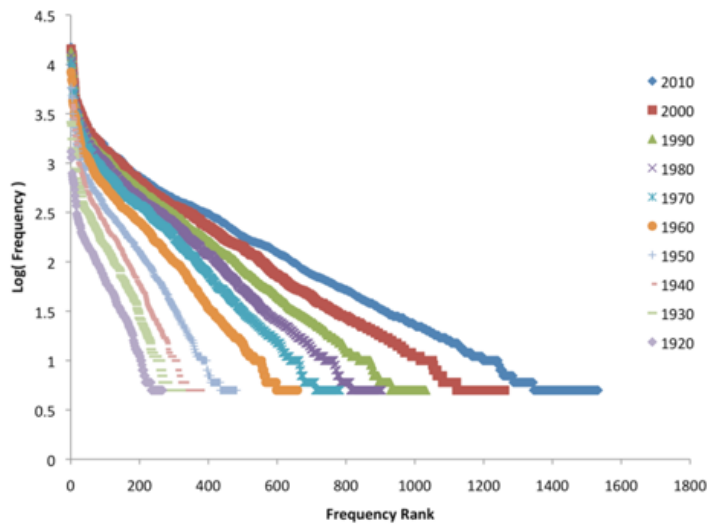

**Figure 8**: Log frequency x frequency rank plot of the cumulative distribution of the names recorded in the social security records (minimum frequency = 5) in Delaware in each decade 1910-2010 (all linear $R^2 > .98$).

Although Delaware's population increased five-fold in the period by 1910-2010 (from around two hundred thousand people to around one million, making it likely that even these data reflect a degree of aggregation), Figures 7 and 8 clearly show that the distribution of first names recorded in the State in each decade was geometric, as was the cumulative distribution in each decade (the average $R^2$ fit for the cumulative distribution of first names to an exponential was = .99 across each decade form 1910-2010, whereas by contrast the average $R^2$ to a power law = .83; t(9)=34.5). It thus follows that these distributions differ markedly from that of the US aggregated as a whole, where name distributions followed a power-law in this period.

The Delaware findings were then replicated in analyses of social security data for Rhode Island (50th for area, 41st for population, and 2nd for population density) and New Hampshire (46th, 43rd, and 41stnd; Figure 9), further supporting the suggestion that local, empirical distributions of American first names are significantly different to the aggregate distribution of all US first names.

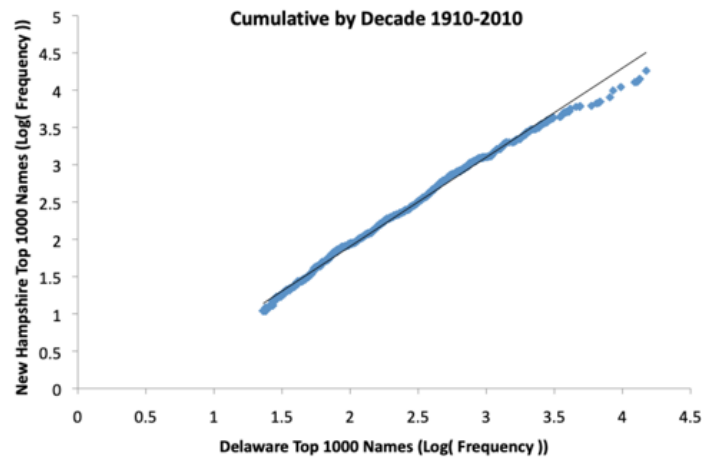

**Figure 9.** Point-wise comparison of the cumulative log frequencies of the 1000 most frequent names given in Delaware and New Hampshire, 1910-2010 ($R^2$=.995).

These results indicate that while the size of the distribution of first names used in English-speaking communities has grown considerably in the past two centuries (increasing name processing costs across the English language as a whole; see Ramscar et al, 2014), when they are disaggregated, the empirical distributions of English first names are still geometric. Moreover, given the historical similarities across name distributions (Lieberson & Lynn, 2003), these findings suggest that empirically, this tendency may be universal when the communicatively relevant distributions of first names in the World's languages are analyzed.

*3.6.5 Names, memorylessness and the source code sampling problem*

Geometric and exponential distributions possess an interesting mathematical property, in that they are both *memoryless* (the exponential distribution is unique in being the only continuous distribution to have this property, just as the geometric distribution is unique in being the only memoryless discrete distribution). The memoryless property is best explained in terms of waiting times: If the probabilities of encountering another individual in a community at any given point in time are distributed exponentially, then because of the way that exponential distributions interact with the laws of conditional probability, it can be proven that the probability of encountering someone at any specific point in time $t_n$ is independent of the time that has elapsed since $t_1$, the time the last person actually arrived.

A counterintuitive result of this proof is that if the waiting periods between encounters with people are exponentially distributed, then the likelihood of encountering another individual in 1 minute is independent of the time that has elapsed since a person was last encountered, such that the likelihood is the same 30 minutes after the last encounter as it was 2 minutes after, and it will remain the same 2 hours later. The same mathematical laws also apply to names: if names are distributed exponentially, then randomly being introduced to an individual named John at any given point in time will have no effect on the likelihood that the next individual one is introduced to will be called John or not. It is also notable that, empirically, there is an abundance of evidence indicating that learning mechanisms have evolved in ways that appear to have optimized their functions towards learning and representing exponential distributions of probabilities (see e.g., Rescorla & Wagner, 1972; Heathcote, Brown & Mewhort, 2000; Trimmer, McNamara, Houston & Marshall, 2012). Taken together, these factors suggest that when a community of learners is exposed to a set of geometrically distributed names, every member of that community will be disposed to learn a model of the probabilities of those names that would be, to all intents and purposes, the same as any other member's model. (The underlying concept was illustrated in the earlier comparison of the distributions of Korean an US Vietnamese names, Figure 10, which indicated that the same rank probability model for first names emerges regardless of whether 1 million or 38 million names are sampled.)

If it is the case that individuals with different sized – and even partial samples – from geometric distributions will be more likely to share probability estimates for their jointly attested items than they would when sampling from other distributions, it follows that when, for example, first names are distributed geometrically, the information value of any given first name in context will be relatively constant across an entire community, regardless of the extent to which any given individual has sampled the name distribution. While the question of whether (or how much) these theoretical considerations shape individual communicative models is ultimately an empirical one, from the perspective of explaining how human communication actually works, then if we assume that human communicative models are shaped in this way (which, empirically, is an open question), it could help explain how people whose individual samples of linguistic codes can be guaranteed to vary enormously (Keuleers et al, 2015; Ramscar, et al, 2017) nevertheless come to share the same (or at least a sufficiently similar) source codes of the kind that <u>all</u> probabilistic theories of human communication assume (whether explicitly or not).

## 4. How do name grammars serve their communicative function?

*"The hard core of information theory is, essentially, a branch of mathematics, a strictly deductive system. A thorough understanding of the mathematical foundation and its communication application is surely a prerequisite to other applications" (Shannon, 1956, p3)*

Some of the most difficult problems faced by the idea of meaning compositionality arise in relation to personal names (Russell, 1919; Searle, 1971; Donnellan, 1972; Boersema, 2000; see Fara, 2015, for a review). By contrast, since it seems clear that the initial tokens in the name systems across the world's languages are distributed geometrically, it follows that name grammars are ideally structured to serve as codes to enable communities of speaker/hearers to communicate information about identities in a way that approximates the kind of communication process described by Shannon (1948). Accordingly, providing a functional account of names in communicative terms is something of a straightforward matter.

To begin with a few preliminary assumptions: First, independent of their linguistic experience, it seems clear that that a neurotypical human brain is well adapted to the task of discriminating individuals based on their specific facial characteristics and other discriminating information (see e.g. Kanwisher McDermott & Chun, 1997); And second, it is equally clear that a neurotypical human brain is capable of learning to tune the representations of individuals that it acquires in this way in order to discriminate between labels that are regularly associated with individuals (Ramscar & Port, 2015). This then raises the question of how people use the associations between representations of individuals and representations of names to communicate, because when it comes to communication, Shannon made clear:

> *"The fundamental problem … is that of <u>reproducing</u> at one point either exactly or approximately a message selected at another point. <u>Frequently the messages have meaning; that is they refer to or are correlated according to some system with certain physical or conceptual entities. These semantic aspects of communication are irrelevant to the engineering problem</u>. The significant aspect is that the actual message is one selected from a set of possible messages. The system must be designed to operate for each possible selection, not just the one which will actually be chosen since this is unknown at the time of design."* (Shannon, 1948, p. 379, my emphasis).

To answer the question raised above, it is important to note that when seen from this perspective, the relationship between an encoding and its meaning is best characterized in terms of a shared code that enables senders and receivers to discriminate between any experiences or states they might wish to communicate (Ramscar & Baayen, 2013). Since in the case of names, these experiences relate to identities, it follows that the communicative needs of a society that wished to communicate about just two identities would be fully satisfied by a code with two signals, 0 and 1. In order to communicate about more identities, this code could then be extended in two ways: by adding codewords (e.g., 3, 4) and/or by combining codewords into strings (11, 01, 1234, etc.).

If one were to merely add codewords to a system, then the problems associated with unique name systems will soon arise. The entropy of the set of names in the code will increase rapidly, making names harder to remember, and the problems associated with code incompleteness will increase, because users of the code will constantly encounter codewords that are unattested in their prior experience. By contrast, if one were to combine the codewords for names into sequences, and restrict the first element in each sequence to an element selected from a small, geometrically distributed set, this would serve to both discriminate names from other signals, and allow the entropy associated with the size of any given set of first names to be minimized. Moreover, in an ideal world, a combinatoric code whose function in communication was that of discriminating between identities would also utilize a set of codewords whose forms both discriminated them as a set from other kinds of signals, as well as making individual name tokens discriminable from one another.

This idea can be easily grasped if we consider a community that wishes to employ more messages to communicate more experiences, such as two states (danger and safety), and two different identities. Encoding these messages as danger = 0000, safety = 0011, and the two identities = 1111 and 1100, will efficiently allow the message a speaker selects to be successfully reconstructed by a hearer. Hearing an initial 11 will enable the hearer to eliminate the possibility that anything other than an identity is being communicated (whereas 00 will signal the opposite), and then hearing either 11 or 00 will enable the hearer to eliminate the identity that is not being communicated (with all of the messages being discriminated maximally according to the degrees of difference afforded by the coding scheme).

It is important to emphasize that this code <u>is not</u> structured to 'encode meanings.' It is simply structured to maximize the discriminability of messages (in this case unique codewords) which can in turn be *correlated* with semantic states. That is, the code is structured so as to best enable a hearer to eliminate all of the other messages that might have been sent, such that the hearer is able to identify (select) the message that is actually being communicated by elimination. If the hearer is also aware of the meanings of the semantic states that are correlated with each codeword, then the same process that allowed them to select the message being communicated will eliminate their uncertainty regarding the speaker's communicative intentions.

Accordingly, when seen from this perspective, Shannon's somewhat notorious comment that "[communicated] messages frequently… have meaning; [however] these semantic aspects of communication are irrelevant to the engineering problem," (Shannon, 1948, p. 379) can be seen to be a clear statement of the relationship between meanings and communicative codes. It does not mean that information theory 'leaves meaning out', as is often claimed, simply because from the perspective described here, meanings are never 'in' codes. Rather, the structure of codes serves to maximize the discriminability of messages, where messages are simply states of the source code (source symbols, or sequences of source symbols). Meanings only enter the picture when 'meaningful messages' are systematically correlated with coded messages, as tends to be the case in natural languages. Critically, in these cases, coded messages <u>do not</u> serve to convey 'meaningful messages.' Instead, in the same way as coded messages are identified by a process of elimination, the 'meaningful messages' speakers intend to send will be identified by a process that eliminates the alternative meanings that might have been intended.

This seemingly perverse and abstract characterization of communication can be put this in terms that are more intuitively graspable as follows: one might imagine that in an eliminative coding system, when a speaker selects an identity to communicate, and then signals it using a name message, the code will be configured to maximize the ability of the hearer to recognize that a name has been signaled. Then, one might expect that this selection and recognition process would be further enhanced by increasing the discriminability of names *vis a vis* other types of speech signals; and one finally might expect the communication process to still further be enhanced by maximizing the discriminability between names themselves.

For English at least, this abstract theoretical characterization captures the properties of first name tokens with remarkable accuracy. The onset of a name in speech is typically signaled by the arrival of an already well-known name token taken from a very small set (in 18[th] Century Scotland or modern Korea, 80% of the identities anyone might wish to communicate will begin with a member of a set of only 20 first name tokens). Patterns of stress and syllabic contrast in English separate names from nouns, while English women's names are longer than men's, are more likely to begin with an unstressed syllable, and are more likely to contain a vowel (Cutler, McQueen, & Robinson, 1990). Further, the forms of most frequent 3 male and 3 female names – *John, William, Thomas, Mary, Elizabeth* and *Anne*, which historically represent 50% of all name

tokens – have remarkably few shared features, which means that this set of forms also serves to minimize mutual acoustic confusability.

At the same time, names such as *John, Will, Tom, Mary, Liz* and *Anne* are meaningful only to the extent that they correlate with identities. It is hopefully clear that nothing in the structure of *John, Will, Tom, Mary, Liz* or *Anne* serves to 'convey' any of the meanings that are correlated with them. *John, Will, Tom, Mary, Liz* or *Anne* simply serve to maximize the likelihood that *Johns* (identities correlated with John) will be easily discriminated from *Wills, Toms, Marys, Lizes, Annes* etc. in communication.

From this perspective, one of the reasons that giving an account of the meanings of names such as *John* has proven to be a problem for compositional theories is that these theories tend to focus on trying to provide an elemental account of the meanings of lexical units in *isolation* (i.e., their focus is on accounting for how individual concepts contribute to the meaning of larger structures). By contrast, arguably one of the most important insights provided by information theory is that codewords contribute information as part of a *system* (i.e., not in isolation): the information value of any given codeword is a function of the set of codewords it belongs to, and it contributes that information in an *eliminative* rather than compositional manner. As Hartley (1928) put it:

> *"In the sentence, 'apples are red,' the first word eliminates other kinds of fruit and all other objects in general. The second directs attention to some property or condition of apples, and the third eliminates other possible colors."* Hartley, (1928, p 536).

Seen from this perspective, the information contributed by the first name *Mary* is not a function of some abstract 'concept of a Mary,' but rather a function of *Mary* belonging to a (highly) structured set of alternatives. When *Mary* occurs in a signal, a hearer now knows that the sets of identities beginning with *John, Will, Tom, Liz, Anne* etc. are not being communicated (see Figure 10).

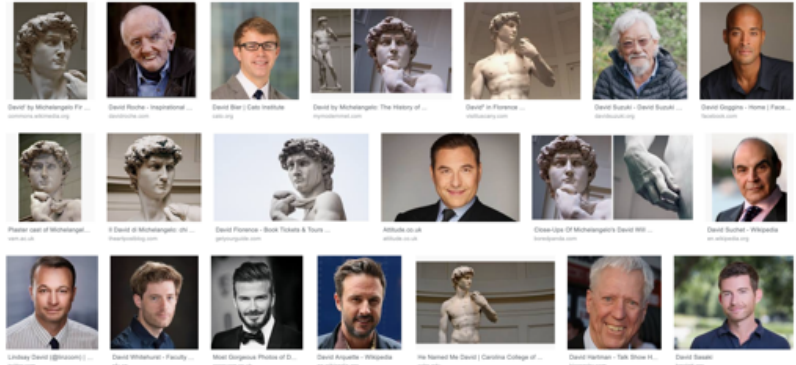

david

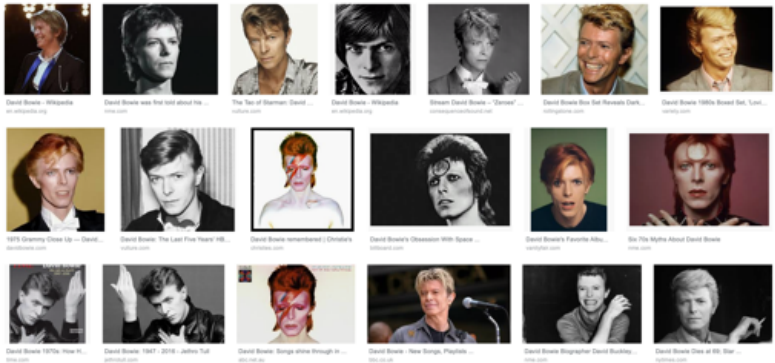

david bowie

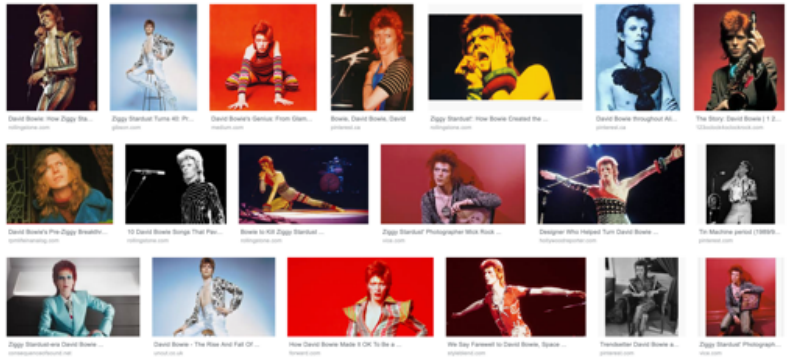

david bowie ziggy period

**Figure 10.** Pictures of the identities that are iteratively discriminated by the search terms "david", "david bowie", and "david bowie ziggy period" by Google image search on 13/2/2019. The search term "david" simply eliminates pictures not related to davids, whereas "david bowie" and "david bowie ziggy period" iteratively focus the search by a process of elimination. This process is entirely consistent with Shannon's (1956, p3) description of information theory as being, "a strictly deductive system." From the perspective described here, names can also be seen as contributing to meaningful communication in the same way: not as part of an 'inductive' process that conveying meaning, but rather as part of a deductive process in which they serve to eliminate alternatives.

Historically, because *John, Will, Tom, Mary, Liz* and *Anne* were well known (they were members of a small set of first names that were readily discriminable from other kinds of lexical signals / parts of speech) their occurrence clearly marked that a name was being signaled. However, in many contexts this communicative function will have been further reinforced by the fact that grammars tend provide information that serves to usefully discriminate between important parts of speech such as names. For example, while English articles obviously serve to discriminate common nouns from proper nouns, it is also the case that the grammar regularly discriminates between personal names and place names:

(5)     John*: Where are you doing this summer?*

      Mary*: I am going to ! Los Angeles.*

      John*: What for?*

      Mary*: I am going to ! see Angela.*

(6)     *I am flying over to ! Los Angeles*

      *I am flying over to ! see Angela.*

(7)     *I am going on holiday with ! Los Angeles.*

      *I am going on holiday with ! Angela.*

(8)     *I am going to ! Los Angeles on holiday.*

      *I am going to ! Angela on holiday.*

Accordingly, in many contexts where *John, Will, Tom, Mary, Liz, Anne* etc. occur, the hearer will be able to deduce by elimination that a name message has been selected, and hence that the speaker intends to communicate an identity, <u>prior</u> to a first name actually being signaled.

It further follows from the distributions of first names that the information that the amount of information that first names will contribute to the selection of an identity in context will vary in a highly structured way. In many contexts, the occurrence of *John, Will, Tom, Mary, Liz, Anne* etc. will be sufficient to eliminate semantic uncertainty in a hearer. That is, in these contexts, the hearer will be able to select the appropriate identity from the first name alone (with the likelihood of this occurring being related to the amount of information any given first name contributes, e.g. *John* versus *Matthew*). In other contexts, the distribution of first names will inevitably mandate that more discriminative information – in the form of further name tokens – must be provided.

*Smith, Jones, Taylor, Brown, Williams, Wilson, Johnson, Davies, Robinson, Wright,*

*Thompson, Evans, Walker, White, Roberts, Green, Hall, Wood, Jackson, Clarke.*

**Table 1.** 20 most frequent English last names in England, 2001 (McElduff, Mateos, Wade, & Borja, 2008).

As Table 1 shows, although the (recent) fossilization of English last names by custom and legislation has tended to both diminish their communicative efficiency and obscure their communicative function, historically the name tokens that followed first names in English tended to be highly predictable. These last name tokens either simply re-sampled from the set of name tokens (i.e., first names were reduplicated as patronyms, e.g., *Jones, Williams, Johnson*), or else re-deployed high-frequency adjectives (*White, Brown*), occupational names (*Smith, Wright*) or names for locations (*Green, Hall, Wood*). Further, the practice of redeploying high frequency lexical tokens from other parts of speech as last names is common across the other languages described above, which means, in other words, that although names serve a *sui generis* function – names essentially lexicalize of the identities that they correlate with, which might in theory require the deployment of *vast* lexical resources – it turns out that, historically, naming practices across the world (embodied in what turn out to be strikingly similar name grammars) have managed to satisfy these functional requirements using only relatively small sets of highly predictable codewords.

Which is also to say that it is clear that the name grammars of the world's languages all appear to provide a remarkable degree of partial attestation for novel forms in a domain that appears, at least in abstract, to pose a particularly serious challenge to human communication. This is because, historically at least, the practice of starting a name with a codeword drawn from a small, well-known set meant that in most contexts, the fact that a first name (and hence a correlated identity) was being communicated would be unambiguously clear to a hearer; and second, the fact that the kinds of words encoding further discriminating information about correlated identities is (or at least, was) highly predictable meant that even where the set of codewords communicating an identity was previously unattested, a hearer would still have a great deal of information about the form of a name signal, such that the hearer's uncertainty about an unattested form would be relatively low. (This of course means that when a name sequence, e.g., *Richard Greenhall* is not correlated with an identity known to the hearer, the hearer at least learns either a) there is an identity correlated with *Richard Greenhall*, and that the hearer might at some point learn to put a face to the name, or else b), that *Richard Greenhall* is not correlated with an actual identity, because the correlated identity might be fictitious like *Sherlock Holmes*, or hypothetical, like *the present King of France*. In either case, the name sequence *Richard Greenhall* will eliminate some uncertainty on the part of the hearer, albeit this reduction in uncertainty will be less than in a case where the intended identity is actually known to the listener.

Finally, if we assume that the empirical distributions of other name tokens are similar to those of first names (as the information sensitivity of naming described above suggests), it follows that the name grammars of the World's languages generate signals that both minimize the average cost of name processing (see also Meylan & Griffiths, 2017), smooth the information communicated by each token in a signal (see also Aylett & Turk, 2004; Fedzechkina, Jaeger & Newport, 2012), as well as making names easier to recall (Dye et al, 2017).

This theoretical analysis name grammars as information systems – i.e., communication systems that are combinatoric but not compositional – is further supported by another surprising empirical fact. As Figure 3 (above) showed, the distribution of popular names in England appears to have been remarkably sensitive to population growth. This suggests that when 19$^{th}$ Century parents chose the names that communicated the identities of their children, they did so in a way that was sensitive to the discriminatory power of those names in a system where last names were fixed, and where the population was growing (i.e., the only point of flexibility in this system is in the selection of first names, and what Figure 3 indicates is that first name choices changed in a way that perfectly reflected to information pressures on the system).

| 1920 | 1940 | 1960 | 1980 | 2000 | 2010 |
|---|---|---|---|---|---|
| Mary | William | John | Michael | Michael | Michael |
| John | John | Robert | Jennifer | Matthew | Matthew |
| William | Robert | James | John | Christopher | Ryan |
| James | James | Michael | Christopher | Tyler | Jacob |
| Helen | Mary | David | James | Joshua | Joshua |
| Margaret | Charles | William | David | Ryan | William |
| Dorothy | Barbara | Mary | Robert | John | Madison |
| Charles | Richard | Richard | William | Andrew | John |
| Joseph | Donald | Deborah | Brian | James | James |
| Elizabeth | Joseph | Linda | Jason | Nicholas | Anthony |
| George | George | Patricia | Matthew | William | Christopher |
| Anna | Betty | Thomas | Joseph | Emily | Emily |
| Robert | Patricia | Susan | Amy | Zachary | Nicholas |
| Mildred | Joan | Joseph | Michelle | Brandon | Andrew |
| Ruth | Margaret | Charles | Daniel | Jessica | Alexander |
| Edward | Shirley | Karen | Kimberly | Ashley | Joseph |
| Catherine | Dorothy | Mark | Thomas | Robert | Olivia |
| Frances | Thomas | Donna | Heather | Sarah | Daniel |
| Thomas | Doris | Barbara | Melissa | Kyle | Ethan |
| Alice | Edward | Sharon | Jeffrey | Joseph | Tyler |

**Table 2**: Changes in popular names in Delaware across the past Century.

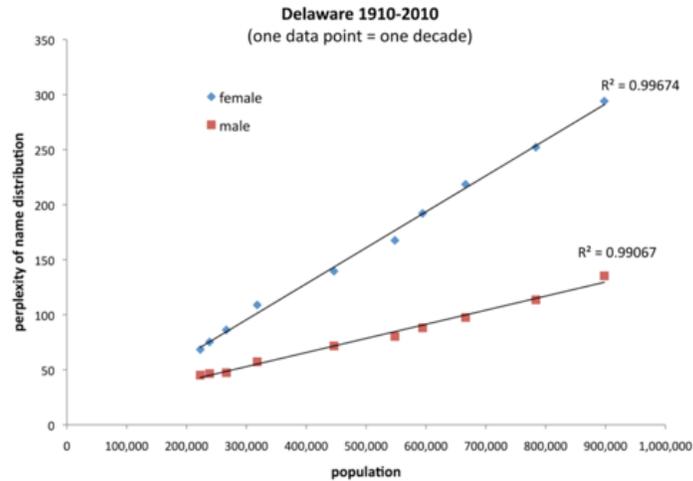

**Figure 12**: The increase in the information perplexity of the cumulative distribution of the names (divided into male and female) recorded in the social security records (minimum name frequency = 5) in Delaware in each decade 1910-2010 plotted against the size of the growing population.

Table 2 shows the most popular names in Delaware at various points in the past Century. When seen in isolation, this might appear to reflect a system in flux, the outcome of a series of individual choices that reflect numerous parents simultaneously attempting to find a unique first name for their child (Gureckis & Goldstone, 2009). However by contrast, Figure 11 which plots the increase in the information perplexity[13] of the cumulative distribution of the male and female names in Delaware against population growth, suggests a process more akin to that observed in 19th Century England, in which changes in the first name distribution closely reflect the informational constraints on the discriminatory power of compound names (comprising a first and last name) in a growing population in which last names are fixed: As can be seen, the increase in the perplexity of first names is almost perfectly correlated with the corresponding decrease in the discriminative power of any given first name as the population grows.

The sensitivity of collective name choices to uncertainty (i.e. information) is further attested to by the difference in the increase in the information of female first names as compared to male first names: whereas the information in virtually all male first/last name compounds are fixed at birth, marriage raises the possibility of changes in the information in female name compounds (*Jane Redbeard* might become *Jane Smith*). Although it is unlikely that this extra uncertainty consciously enters the minds of parents, it is near-perfectly reflected in their cumulative naming choices, as reflected by the fact that the information in female names has increased more sharply with population growth as compared to male first names, a finding that further highlights how parents appear to choose names that maximize discriminability and minimize information costs within <u>all of</u> the constraints that have been imposed by the overall system.

---

[13] Because entropy values are a function of the way items are distributed — depending on their distribution, many or few discrete items might have an entropy of 4 bits — information entropy can be a difficult idea to intuitively grasp. A solution to this problem in computational linguistics is to convert entropy into perplexity (Bahl, Baker, Jelinek, & Mercer, 1977), which expresses bit values in terms of a distribution of equally likely outcomes, calculated as $2^H$ (so that a distribution with an entropy H = 3 bits has a perplexity of 8). Information entropy is non-linear (logarithmic), whereas the transform to perplexity provides a linear measure more suitable for the comparison in Figure 12.

### 5. The statistical structure of "semantic domains"

Information theory successfully recast many of the problems posed by electronic communication by framing them in terms of discriminating an actual message being transmitted from the total space of messages that could possibly have been transmitted (Shannon, 1948, 1956), and the foregoing indicates that the function of names can be successfully revealed by recasting their communicative contribution along similar, discriminative, lines. By contrast, other lexical items have long been thought to communicate compositional meanings, as well as being thought to be Zipf distributed. This raises an intriguing question: Do name grammars represent a functionally distinct, perhaps evolutionarily older, linguistic system – as the neuropsychological processing differences between names and other words might suggest (see e.g. Müller & Kutas, 1996; Proverbio, Lilli, Semenza & Zani, 2001) – or is all of human communication actually based on discriminative principles?

One reason to doubt that that name grammars are alone in being information systems – i.e., that natural language grammars may not in fact be compositional systems – comes from research that has shown that many aspects of "word meanings" can be captured by measures of the similarities and dissimilarities in the relationships between words in large text corpora (e.g., Landauer & Dumais, 1997; empirically, these measures have been shown to fit well with human judgments and behavior across a range of tasks associated with semantic processing, McDonald & Ramscar, 2001; Jones & Mewhort, 2007; Johns & Jones, 2010; Ramscar et al, 2010 a,b; Ramscar, Dye & Klein, 2013; Baayen, Milin & Ramscar, 2016). Models of 'distributional semantics' – which discriminate between place names and personal names as a matter of routine (e.g., Yarlett, 2008) – usually take their inspiration from the Distributional Hypothesis (Firth, 1957), which posits that many aspects of a word's 'meaning' can determined from the linguistic contexts in which it occurs. As such, distributional metrics are typically described at the lexical level, as capturing aspects of the meaning of individual lexical items, albeit that it is unclear that the conventional understanding of 'meaning' in relation to co-occurrence models actually makes sense (Ramscar, Dye & Hübner, 2013). What is important to note for present purposes is exactly what it is that 'distributional models of semantics' actually measure, which is the conditioning history of a word in relation to the other words in a language. For example, in a model that measures distributional similarities by building a 'semantic vector' based on the distributional co-occurrence patterns of a word in relation to a set of other words, the content of that vector is simply a record of the input that someone exposed to that distribution would receive as they learn to condition their expectations about each word's behavior in relation to other words.

Because learning – including classical conditioning – is best characterized as a predictive, discriminative process (Ramscar et al, 2010; Ramscar & Port, 2016), it follows that if two or more words have the same conditioning histories (that is, if the vectors of their co-occurrence patterns in relation other words are identical), then while someone exposed to this distribution will learn to discriminate these words from the words in the rest of the lexicon that don't share the same conditioning history, she won't learn to discriminate them from one another. Further, because learning is a probabilistic process (i.e., the degree to which learners will come to discriminate the expected behavior of one word from that of another will be a matter of degree) where two or more words have conditioning histories that vary only slightly from one another, but greatly from other words, a learner's expectations about the behavior of the words in such a set will be far less discriminated within set than they will be from the rest of the lexicon, such that members of the set will tend to cluster in the lexicon.

These considerations suggest that when it comes to explaining what models of 'distributional semantics' actually capture, a discriminative perspective can shed far more light on their relevance to linguistic theory than the compositional ideas that dominate current thinking. From a discriminative perspective, the fact that these models capture many semantic similarities between words does not reflect the fact that individual word 'meanings' are (somehow) the product of a word's relationships with other words. Rather, just as systematic patterns of variance in English grammar discriminate 'ontologically' between personal names and place names (5-8, above), distributional models are able to capture many aspect of the semantic similarities between words because the grammar of English actually encodes a very rich functional ontology that enables (and in fact, causes) learners to learn to discriminate sets of words that are meaningfully related to one another from other words that are not, and then to use these on

This means in turn that prior to a name being heard, a speech signal that is structured in this way will already have communicated some semantic information to a hearer, simply because she will now know that a name is likely to occur. Further, in the same way that the grammar of English often provides a hearer with information that leads her to expect a name prior to its actually occurring (thereby providing her with semantic information), so other systematic patterns of variance in the linguistic distribution can provide hearers with semantic information in context. Although this semantic information is usually thought to be 'carried,' or 'denoted,' by content words, in such contexts some of this content will be communicated prior to what are normally thought of as 'content words' actually being encountered.

Importantly for current purposes, if the grammar does encode this kind of ontology, and if the rest of human communication does in fact work along the same lines as it seems names work, then it follows that human communication is not a process in which content words 'carry' meanings, and in which semantics are 'compositional' in the sense that utterance meanings additively comprise the meanings of content words (modified in some way by order and the presence of function words). Instead, what we end up with is a very different view of human communication to the one that has historically been supposed. From this perspective, the reason why distributional models capture so much semantic information is not because they capture 'aspects of individual word meanings.' Rather, because sets of words with similar conditioning histories are less discriminated from one another in learning than they are from the rest of the lexicon, it thereby follows that where semantically similar words co-vary systematically in the lexicon, the lack of discrimination produced by this invariance will 'capture' those similarities, thereby encoding these coarse semantic similarities in signals (and blurring distinctions between 'content' and 'function' words). If this is the case, then it suggests that all of human communication may actually involve the same kind of discriminative processing as names. That is, rather than being compositional, all meaningful linguistic communication may involve the reduction of semantic uncertainty in the same way the semantic processing of names does.

Critically from this perspective, although linguistic signaling can, to some degree, be though of in combinatorial terms (in that signaling makes use of sequences of discrete form contrasts), the process that hearers use to reconstruct linguistic messages (i.e., meanings) is discriminative. Hearers do not 'extract' meanings that are compositionally 'encoded' in signals, as linguists have long supposed (albeit in ways that vary from vagueness to incoherence; see Ramscar & Port, 2015), but rather, a hearer uses a signal to help her discriminate the message a speaker intends them to receive from the messages a speaker might have sent (which is exactly how, Shannon, 1949, defines 'communication'); And what makes this communicative process possible is the fact that the speaker and listener both share a richly structured common code that systematically patterns forms with experience.

The discriminative account of communicative processing outlined here makes a clear, falsifiable prediction: If what distributional models of semantics actually capture is the kind of grammatically coded ontology just described, and if the variance classes encoded by this ontology contribute to meaningful communication in the same way that the variance that discriminates personal names as a class does, then it follows that the sets of words discriminated by this kind of grammatical variance must be empirical distributions (the distributions language users anticipate in moment to moment language processing). If this is the case, then for the same functional reasons that that name distributions are geometrically distributed, it follows that words in the variance classes captured by distributional models ought to be distributed geometrically as well.

**5.1 Noun systems as information systems**

The empirical distributions that are discriminated by co-occurrence patterns in child-focused speech are particularly salient to language learning. If these empirical distributions are geometric (i.e., memoryless), this would help explain how it is that the communicative expectations that develop out of a child's interactions with other speakers come to align with the common set of expectations that comprise the source code for communication in her community. Accordingly, in order to examine the degree to which the distributional structure of the linguistic input children are exposed to does discriminate these distributions (and the kind of natural ontology described above), the distributions in a test set of noun clusters that have been shown to be learnable from a corpus of child / parent speech (CHILDES; MacWhinney, 2000) by a broad class of distributional models (Asr, Willits & Jones, 2016; see also Jones, Willits & Dennis, 2015) were examined.

Asr, Willits & Jones (2016) defined this test set using 1,244 high frequency nouns (taken from the CHILDES corpus) that form 30 relatively unambiguous categories (e.g., *bird, body kitchen, mammal, clothing*, etc.), and showed that a wide range of models could learn to discriminate the members of these categories appropriately from the distributional properties of the language children up to age five were exposed to in CHILDES. For example, Table 3 shows the set of nouns in the *clothing* category and Table 4 the set of nouns in the *mammals* category.

*Clothes, Dress, Suit, Shirt, Coat, Hat, Tie, Jacket, Cap, Belt, Uniform, Hood, Shoe, Skirt, Cape, Purse, Boot, T-Shirt, Shorts, Helmet, Outfit, Sweater, Glove, Gown, Underwear, Robe, Sunglasses, Scarf, Blouse, Vest, Bra, Apron, Buckle, Sock, Diaper, Slacks, Sweatshirt, Nightgown, Tights, Bathrobe, Pant, Bonnet, Sneaker, Slipper, Pajama, Sandal, Undershirt, Bib, Mitten, Snowsuit, Shoelace, Jammie,\* P-J\**

**Table 3**: Nouns in the *clothing* category in Asr, Willits & Jones (2016).

*Armadillo, Baboon, Badger, Bear, Beaver, Buck, Buffalo, Bull, Bunny, Calf, Camel, Cat, Cheetah, Chimpanzee, Chipmunk, Collie, Cow, Coyote, Deer, Dingo, Dog, Dolphin, Donkey, Elephant, Fox, Giraffe, Goat, Gorilla, Groundhog, Hamster, Hare, Hedgehog, Hippo, Horse, Hyena, Jaguar, Kangaroo, Kitten, Koala, Lamb, Leopard, Lion, Mammal, Mammoth, Mole, Monkey, Moose, Mouse, Mule, Opossum, Orangutan, Otter, Ox, Panda, Panther, Pig, Piglet, Pony, Porcupine, Pup, Rabbit, Raccoon, Rat, Reindeer, Rhino, Seal, Sheep, Skunk, Squirrel, Steer, Tiger, Walrus, Weasel, Whale, Wolf, Zebra*

**Table 4**: Nouns in the *mammals* category in Asr, Willits & Jones (2016).

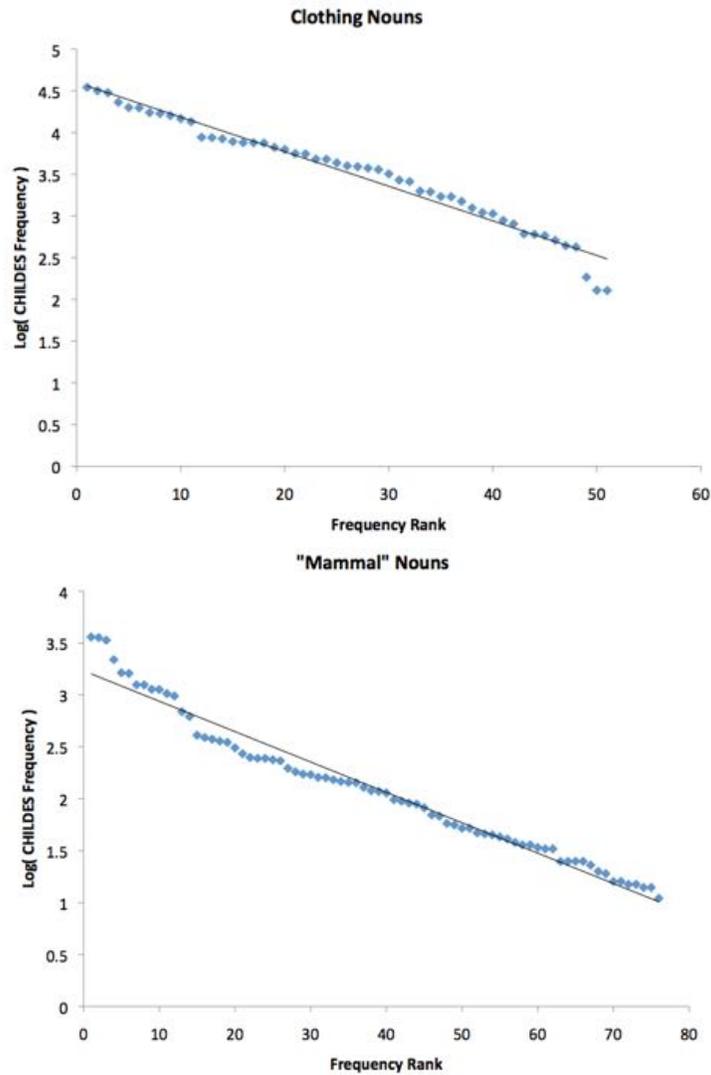

**Figure 12**: Log frequency x frequency rank plots of the *clothing* and *mammal* noun categories.

The empirical frequency distribution of the nouns in each class was examined by extracting the frequencies of each of its members from CHILDES itself. Averaging across all

20 classes, the fits between the items in the empirical distributions taken from child-centered speech were considerably closer to geometric distributions (mean $R^2$= 0.95) than they were to power law distributions ($R^2$= 0.89, t = -3.793); see Figure 12).

**5.2 Verb alternations and communication**

The idea that the distributional variations that make up the grammar of English discriminate systematically within supposedly basic linguistic categories such as verbs has a long history (albeit, historically, this idea has been explored from a very different theoretical perspective). As (9) – (12) illustrate, apparently verbs similar such as please and like are not simply interchangeable across the grammar:

(9)     John ! liked that Bill was always punctual

*versus*

John ! pleased that Bill was always punctual

(10)    John was ! pleased by the fact that Bill was always punctual

*versus*

John was ! liked by the fact that Bill was always punctual

(11)    John ! loved Jill's winning personality

*versus*

John ! enthused Jill's winning personality

(12)    John often ! enthused about Jill's winning personality

*versus*

John often! loved about Jill's winning personality

Detailed descriptions of many of these alternation patterns – which often share semantic similarities in English – have been made (Levin, 1993), along with the classes they form. For example, Table 5 lists the verbs that comprise the "throw" alternation class (from Levin, 1993).

*bash, bat, bunt, cast, catapult, chuck, fire, flick, fling, flip, hit,*

*hurl, kick, knock, lob, loft, nudge, pass, pitch, punt, shoot, shove,*

*slam, slap, sling, smash, tap, throw, tip, toss*

**Table 5**: The "throw" verbs, from Levin (1993).

To explore the distributional properties of this alternation class, the frequencies of its members were extracted from the Corpus of Contemporary American English (COCA, Davies 2009). Figure 13 plots the frequency distribution of the verbs in terms of their log frequency by their frequency rank. As it clearly shows, the verbs in the "throw" alternation class are geometrically distributed.

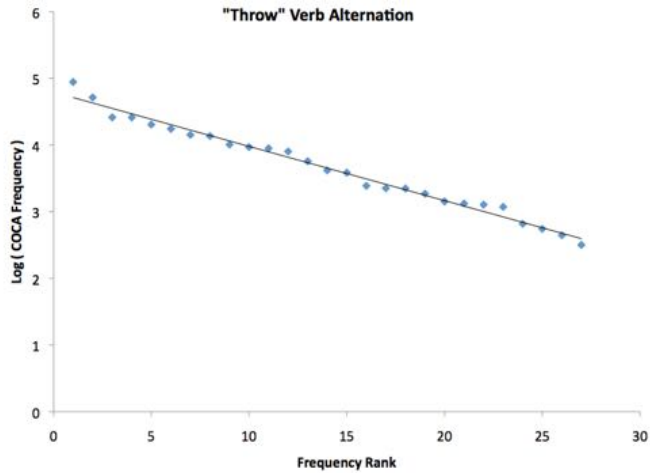

**Figure 13**: Log frequency x frequency rank plot (exponential is linear) of the "throw" verbs ($R^2$=.98).

The results of this analysis are thus highly consistent with the results observed for names and nouns. To further examine this trend, the distributions of a further three sets of alternations from Levin (1993) sharing a broad semantic theme – the "appear," "skate" and "fill" alternations (Tables 6, 7 & 8) – along with two alternations defined in terms of their distributional properties: the locative preposition drop alternation (Table 9) and the "with" preposition drop alternation (Table 10).

*appear, arise, awake, awaken, break, burst, come, dawn, derive,*

*develop, emanate, emerge, erupt, evolve, exude, flow, form, grow, gush,*

*issue, materialize, open, plop, pop up, result, rise, show up, spill,*

*spread, steal, stem, stream, supervene\*, surge, turn up, wax*

**Table 6**: The "appear" verbs, from Levin (1993). *Supervene was excluded from further analysis because of its low COCA frequency.

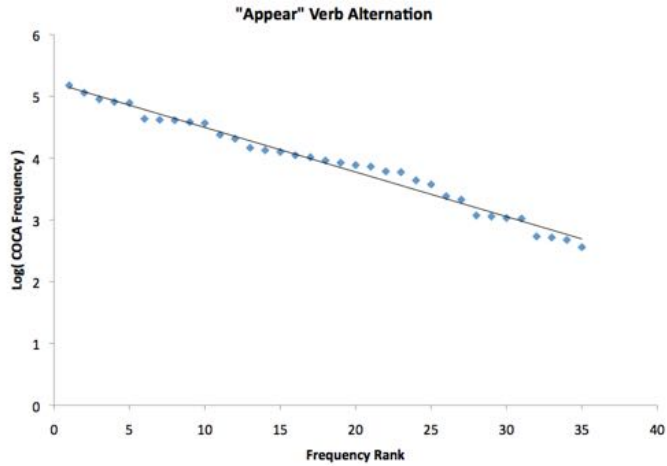

**Figure 14**: Log frequency x frequency rank plot of the "appear" verbs ($R^2=.98$).

*balloon, bicycle, bike, boat, bobsled, bus, cab, canoe, caravan, chariot, coach, cycle, dogsled, ferry, gondola, helicopter, jeep, jet, kayak, moped, motor, motorbike, motorcycle, parachute, punt, raft, rickshaw, rocket, skate, skateboard, ski, sled, sledge, sleigh, taxi, toboggan, tram, trolley, yacht.*

**Table 7**: The "skate" verbs, from Levin (1993).

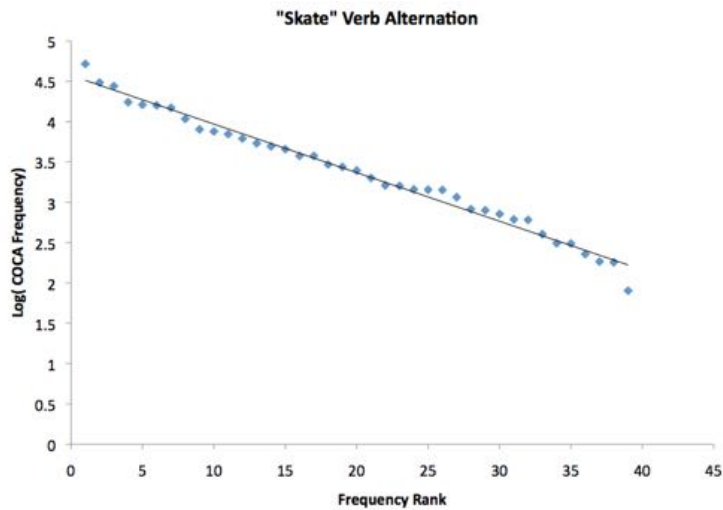

**Figure 15**: Log frequency x frequency rank plot of the "skate" verbs ($R^2=.98$).

*adorn, anoint, bandage, bathe, bestrew, bind, blanket, block, blot bombard, carpet, choke, cloak, clog, clutter, coat, contaminate cover, dam, dapple, deck, decorate, deluge, dirty, dot, douse, drench edge, embellish, emblazon, encircle, encrust, endow, enrich entangle, face, festoon, fill, fleck, flood, frame, garland, garnish imbue, impregnate, infect, inlay, interlace, interlard, interleave intersperse, interweave, inundate, lard, lash, line, litter, mask mottle, ornament, pad, pave, plate, plug, pollute, replenish repopulate, riddle, ring, ripple, robe, saturate, season, shroud smother, soak, soil, speckle, splotch, spot, staff, stain, stipple stop up, stud, suffuse, surround, swaddle, swathe, taint, tile, trim*

**Table 8**: The "fill" verbs, from Levin (1993).

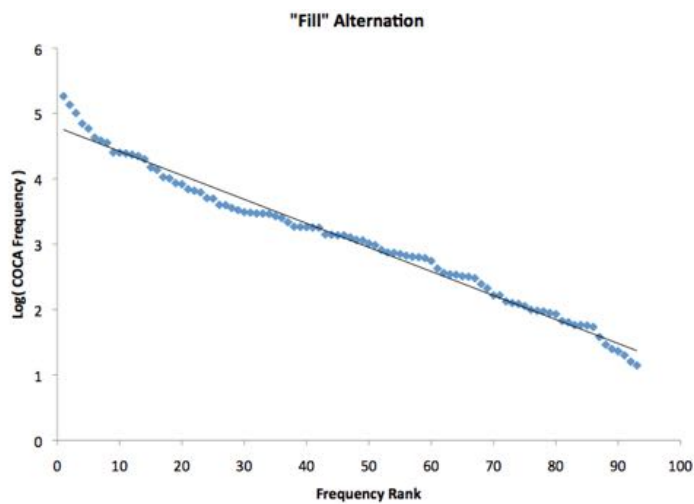

**Figure 16**: Log frequency x frequency rank plot of the "fill" verbs ($R^2$=.98).

*arrive, ascend, bicycle, bike, bounce, canoe, canter, climb, coil come, cross, depart, descend, drift, drop, escape, flee, float, fly gallop, glide, go, hike, jeep, jog, jump, leap, leave, move, prowl, raft ramble, return, revolve, ride, roam, roll, rotate, rove, row, run, sail shoot, skate, ski, slide, spin, stroll, swim, swing, traipse, tramp, travel, trudge, turn, twirl, twist, vault, wade, walk, wander, whirl wind*

**Table 9:** The "locative preposition drop alternation" verbs, from Levin (1993).

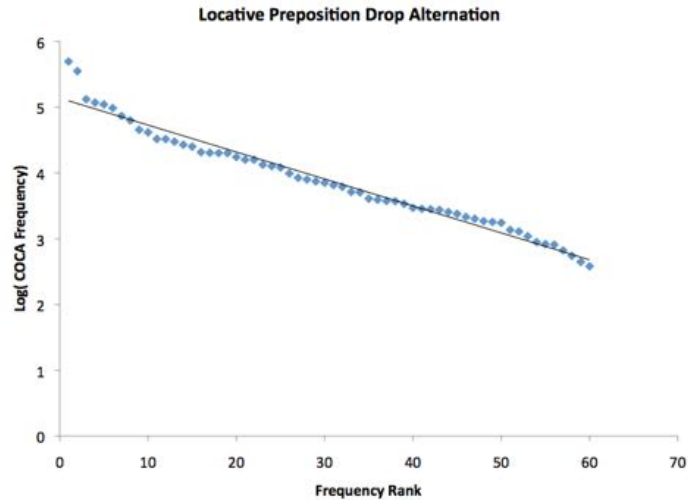

**Figure 17**: Log frequency x frequency rank plot of the "locative preposition drop alternation" verbs ($R^2$=.97).

*agree, argue, banter, bargain, battle, bicker, box, brawl, chat, chatter, chitchat, clash, coexist, collaborate, collide, combat, commiserate, communicate, compete, concur, confabulate\*, confer, conflict, consort, consult, converse, cooperate, correspond, court, cuddle, date, debate, dicker, differ, disagree, dispute, dissent, divorce, duel, elope, embrace, feud, fight, flirt, gab, gossip, haggle, hobnob, hug, jest, joke, joust, kiss, marry, mate, meet, mingle, mix, neck, negotiate, nuzzle, pair, pass, pet, play, plot, quarrel, quibble, rap, rendezvous, schmooze, scuffle, skirmish, spar, spat, speak, spoon, squabble, struggle, talk, tilt, tussle, vie, visit, war, wrangle, wrestle, yak*

**Table 10**: The "with preposition drop alternation" verbs, from Levin (1993). *Confabulate was excluded from further analysis because of its low COCA frequency.

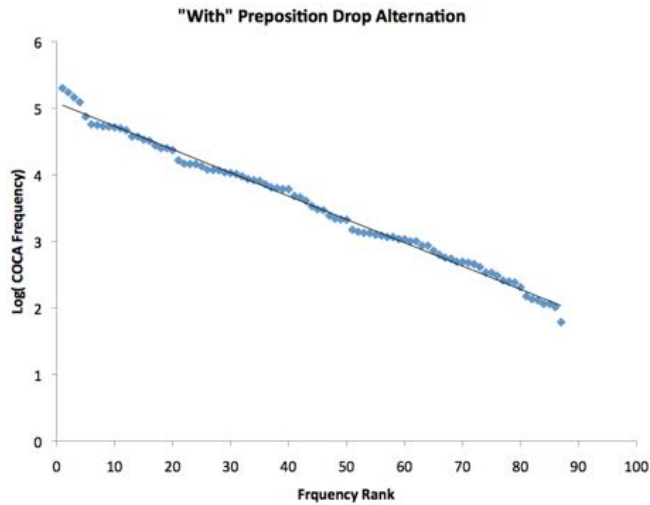

**Figure 18**: Log frequency x frequency rank plot of the "with preposition drop alternation" verbs ($R^2$=.99).

As Figures 13-18 show, in every case, the distribution of these items is geometric (and thus memoryless). This suggests that, as with the empirical distributions of nouns, the distributions of these verbs appears to be optimized to support the same kind of discriminative communication process as names.

## 7. Systematicity in communication

Thus far we have seen how the structure of empirical distributions of words in English (and, as far as personal names go, other languages) support a discriminative communication system. Given the properties of these distributions, it seems likely that the level of systematic informative variation they support goes beyond the orthographic, lexical and distributional levels that discussed so far. Indeed, it is likely that the presence of informative variance / contrasts at the sublexical level can enable us to explain just how it is that the comprehension of rapid conversational speech is possible in the first place.

The idea of informative sublexical variation is best illustrated by example. Consider the speech forms represented by the orthographic form "has" in (13) (14) and (15):

(13)    *Who has John been seeing?*

(14)    *He has been seeing Jack's wife.*

(15)    *He has to stop.*

In the conversational speech of native English speakers the gestures corresponding to these three instances of "has" are not merely readily discriminable, but examples such as this are often used to illustrate the "problems" that "speech perception" poses, especially to children learning language: *how is a learner supposed to discover that these three gestures are tokens of the same type?* Further, and somewhat paradoxically, because linguists also assume that children treat these three gestures are tokens of the same type, this then leads to a further problem, that of "wh-question" forming.

(16)    *John has been seeing Jack's wife.*

If we take a declarative sentence like (16), then the corresponding way of requesting the information contained in this sentence (the identity of the person John has been seeing is that given in (13), and not, as a supposedly naïve learner is assumed to suppose:

(17)    * *Who John has been seeing?*

Because linguists have assumed that linguistic signals are assembled out of basic units of form-meaning mappings (the compositional assumption), and because "has" is assumed to be one of these basic form-meaning mappings, it has been supposed that a child (a 'naïve learner') is faced with near-insuperable problems when it comes to learning the rules for turning a declarative sentence into a question.

There are so many problems with these assumptions that I cannot hope to do justice to them here, although I would remiss not to at least note that studies of child-directed speech have shown that questions represent *by far* the largest single category of utterance that a typical child encounters (Cameron-Faulkner, Lieven, & Tomasello, 2003), which suggests that empirically the assumptions and intuitions that give rise to the "problem of wh-movement" are fundamentally wrong.

What is important to note is that, from a discriminative perspective, "problems" like "wh-movement" <u>do not</u> represent 'deep questions' that need to be answered in order to explain how languages are learned. Instead these questions arise out of theory internal difficulties that arise themselves as a result of over-abstraction. The assumptions that shape way that the "problem of wh-movement" is posed, and the vague and – at this point, naïve – assumptions about learning that underpin it are a direct result of the way that researchers have habitually ignored the <u>many</u> levels of informative variation that are actually available to (discriminative) learners. One such level – already described in relation to empirical distributions of nouns and verbs – is informative variance in the grammar. To state the obvious, the meanings communicated by questioning and declarative utterances are different. It therefore ought to come as no surprise that in a natural communication system that has evolved across communities of learners to maximize informative contrasts (and has also been shaped by the requirement that it be learnable by new additions to these communities), has come to organize its grammar so that distinctions as fundamental to communication as questions and declarations are highly discriminable.

From a discrimination learning perspective, this kind of systematic variation does not make language "unlearnable" for a child (as has often been claimed), but rather it facilitates the learning of the different constructions that are employed in English questions and declarations in exactly the same way that the empirical distributions of words will enable children to learn the many aspects of the semantic ontology of English described above. Accordingly, the systematic structural variation in the grammar of questions and declarations can be seen as being no different to the structural variation in the grammar of place names and personal names. Each serves to discriminate a subset of messages for the purposes of a specific kind of communication. Similarly, by ignoring structural variation in the grammar, and by over-abstracting the notion of a word (i.e., in treating the systematic variance in the spoken forms of "has" as noise rather than signal), traditional approaches to linguistics are blind to the role that informative variation within word forms (i.e., at the sublexical level) is likely to play in making speech possible (at least, in the empirical forms in which it is actually encountered). Indeed, an ever-growing body of evidence supports that suggestion that speech variation is far more systematic and informative than has previously been supposed.

For example, it has been found that speakers produce nouns with longer mean durations when they are pronounced as singulars than as the stem of the corresponding plural (Baayen, McQueen, Dijkstra & Schreuder, 2003), and that listeners are sensitive to the way these different gestures discriminate between the plural and singular version of what it usually thought of as the "same" form. If the segment (the inflection) that is usually thought mark plurality a Dutch noun as a plural is spliced out of a recording, native speakers do not judge the resulting "unmarked" noun to be singular, but rather judge it to be a distorted plural (Kemps, Ernestus, Schreuder & Baayen, 2005). Indeed, listeners' sensitivity to this contrast is such that it causes them to 'perceive' the presence of the suffixes that normally follow then even when they are in fact partly and even completely missing in the reduced acoustic signals that they hear (Kemps, Ernestus, Schreuder & Baayen, 2004).

This highlights a critically important aspect of 'speech perception': what listeners 'hear' when listening to natural speech is not only a function of the sounds speakers emit. Listener's expectations about what they are hearing play a critical role in the process of speech understanding (for review, see McQueen & Cutler, 2012). Thus for example, when the highly 'reduced' forms of frequent words that are actually produced in speech – and which are easily processed in natural speech understanding – are presented out of context, listeners are often completely unable to recognize them.

For example, Ernestus, Baayen & Schreuder (2002) presented native Dutch speakers with a speech sample resembling [εik] (a reduced form of the Dutch word *eigenlijk*; "in fact") and asked them report exactly what they heard. When listeners heard this audio clip presented in its full context, they reported hearing *eigenlijk,* where when they encountered the exact same acoustic signal without context, they reported hearing the meaningless *eik*.

Since speech understanding is a product of both what speakers do and what listeners expect, it follows that it will work best when the actions of speakers and the expectations of listeners are aligned. That is, the efficiency – and even the very possibility – of speech understanding *depends* on the <u>alignment</u> between what speakers do and what listeners expect them to do. For example, Jurafsky and colleagues (Bell, et al, 2009; Jurafsky, Bell, Gregory & Raymond, 2001) have shown that the spoken durations of English function words vary systematically as a function of both the degree to which they can be expected in context, and the degree to which they make following material more predictable. From the perspective described here, these findings indicate that the distribution of the tokens of any given gesture (i.e., word form) within an empirical distribution of words provides yet another level of discriminative structure in language.

To illustrate how this works, consider the processing of *David* in (18):

(18)    John:   *Hi Steve. I'd like you to meet my friend David Smith*

        versus

        John:   *Hi Steve. I'd like you to meet my friend David Brezhnev*

Since it appears that the distribution of first names in a community will inevitably have a geometric (memoryless) distribution, let us assume that as long as Steve and John learned from the same name distribution, *David* will have the same information value in the two models that shape their respective communicative behavior. Accordingly, given that the duration of *friend* in John's speech will be partly determined by the information in the following gesture (Bell, et al, 2009), and given that Steve and John's models of that information will be closely aligned, Steve will be able to make use of the durational information in *friend* in order to reduce his uncertainty about what is about to occur (i.e., Steve will be able to infer from the duration of *friend* whether a low or high information name is about to follow). Similarly, because the duration of *David* in John's speech will be partly determined by the information in next gesture, Steve will be able to make use of the durational information in *David* to reduce his uncertainty about what follows.

Because contexts invariably provide information, these same points will apply whatever John is planning to say:

(19) *The first decade and a half of the twenty-first century was later described as the golden era of linguistic and psychological enquiry*

   versus

   *The first decade and a half of the twenty-first century was later described as the Brezhnev era of linguistic and psychological enquiry*

In both cases in (19), the likelihood of the form or forms that follow *described as* will be communicated to some degree by the shape and duration of the gesture that is used to articulate *the,* because the shape and duration of this gesture will be consistently influenced by the probability of the following form or forms (Bell, et al, 2009; Jurafsky et al, 2001; see also Hall, Hume, Jaeger & Wedel, 2018; Wedel, Nelson & Sharp, 2018). Or, in other words, while my focus here has been on the distribution of lexical forms, and the remarkable way in which these distributions appear to have been adapted by the behavior of generations of learners to facilitate human linguistic communication, it seems clear that the idea of inventories of lexical 'items' is itself an abstraction.

To further illustrate this point, Figure 19 plots some results from a series of analyses by Linke and Ramscar (in prep) of the distributions of the word initial phonetic labels assigned to the observed forms in the Buckeye Corpus of conversational speech (Pitt, Johnson, Hume et al, 2005). The corpus contains 300000 orthographically transcribed and phonetically labeled words from 40 interviews with talkers from Columbus, Ohio. The label set comprises 41 standard aligner phonetic labels, plus labels extended by manner of articulation (nasalized vowels, nasal flaps, flaps, glottal stops and r-colored vowels). These analyses reveal that the overall distributions of phonetic contrasts show better fits to geometric distributions, and moreover, this is also the case when position and syntactic function (defined in terms of part of speech) are taken into consideration, such that the phonetic contrasts at the beginning of words in each syntactic category show a better fit to a geometric than a power law distribution (Figure 19).

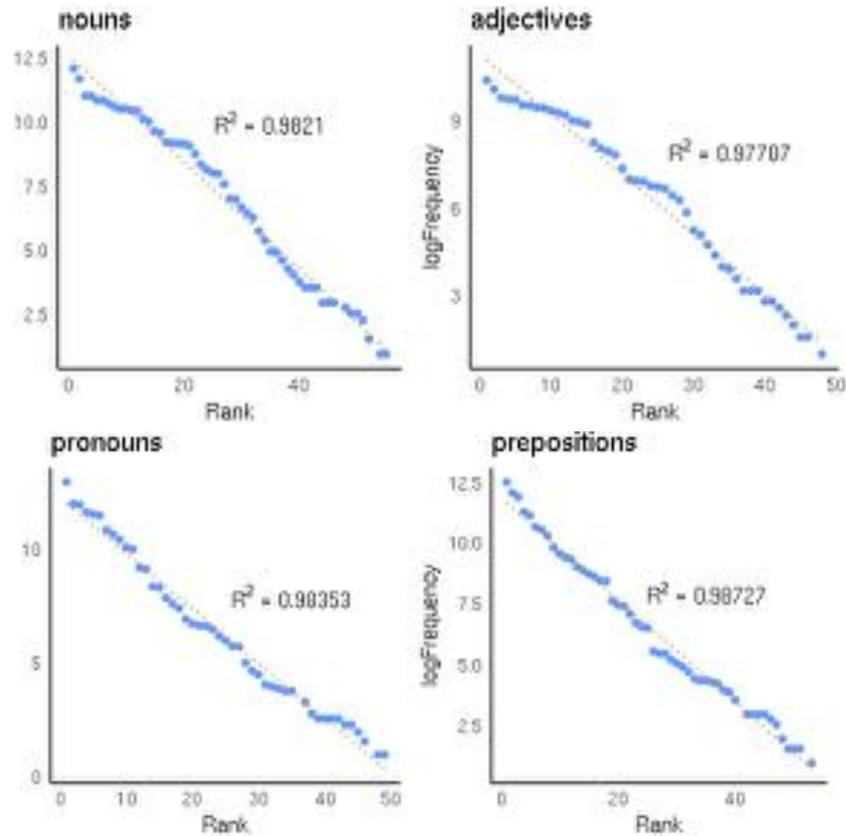

**Figure 19**: Log frequency x frequency rank plots of the distributions of word initial phonetic labels assigned to observed forms across parts of speech in the Buckeye Corpus of conversational speech (Pitt, Johnson, Hume et al, 2005)

Not only do these findings emphasize the idea that lexical "items" are themselves parts of systems, they also hopefully serve to illustrate how systematicity is present in human communicative codes at many (if not all) levels of abstraction.

# 8. Making sense of meaning

*"The belief that words have a meaning of their own account is a relic of primitive word magic, and it is still a part of the air we breathe in nearly every discussion."*
Ogden & Richards (1923)

## *8.1 Moving beyond compositionality*

As I noted at the outset of this article, belief in compositionality – the doctrine that meaningful messages are built up out of smaller units of meaning (etc.) – has dominated and continues to dominate theories of in human communication. Indeed, at times is seems that most linguists (and other researchers in the brain and cognitive sciences) are unable to conceive of human communication in anything other than in compositional terms. As a result, it seems to fair to say that on one hand, even sensible attempts to dispense with compositionality (despite its myriad problems) are generally greeted with incredulity by the research community, whereas on the other hand, even the most outlandish suggestions tend to be given credence, just as long as they conform to the prevailing orthodoxy. For example, the problems posed by compositionality have led leading theorists to the absurdist claim that units of meaning—such as *dog, appear, carburetor, bureaucrat*, and, presumably *entropy, perplexity John, William, Mary* and *Anne*—must be (somehow) specified in the genome:

> *there is good reason to suppose that the [nativist] argument is at least in substantial measure correct even for such words as carburetor and bureaucrat, which, in fact, pose the familiar problem of poverty of stimulus if we attend carefully to the enormous gap between what we know and the evidence on the basis of which we know it. The same is often true of technical terms of science and mathematics, and it surely appears to be the case for the terms of ordinary discourse. However surprising the conclusion may be that nature has provided us with an innate stock of concepts, and that the child's task is to discover their labels, the empirical facts appear to leave open few other possibilities. (Chomsky, 2000, pp. 65–66)*

While many researchers who subscribe to compositionality would balk at endorsing this claim, it hardly matters, first because the idea of *innate concepts* is so completely vague as to contribute virtually nothing to our understanding of human communication, and second because it is clear that no better alternative account of meaning compositionality is on offer.

As I noted at the outset, after half a century of motivated effort, researchers have singularly failed to come up anything approximating a half-coherent empirical account of what a 'unit of meaning' is supposed to be (Ramscar & Port, 2015), and philosophical analyses that have long suggested that 'meaning units' are a fundamentally misguided idea (Wittgenstein, 1953; Quine, 1960; see also Fodor, 1998). Which is to say that although most linguists (and other researchers in the brain and cognitive sciences) clearly believe in compositionality, theoretical accounts of compositional meaning themselves offer nothing beyond blind faith, vagueness, and / or mysticism.[14]

By contrast, when taken together with well-established principles from information theory, the data described here show that for personal names, the empirical facts emphatically open up possibilities for explaining human communication that seem far more helpful than hopeful appeals to innate concepts. Empirically it seems clear that name grammars are perfectly adapted to supporting a discriminative process in which speaker's signals – structured sequences of name tokens – simply serve to eliminate semantic uncertainty in a hearer. These combinations / sequences of name tokens are not compositional, nor do they 'refer to concepts' (whatever concepts are supposed to be). Rather, tokens (and token sequences) eliminatively narrow the space of possibilities within the set of identities that is correlated with them.

At which point, it seems worth re-emphasizing that accounting for the workings of personal names has traditionally proven to be a stumbling block for compositional theories, to the extent that mainstream linguistic theories typically either ignore names altogether, or else relegate them to an afterthought. By contrast, names appear to be an aspect of human communication that is readily explained in discriminative terms.[15] The reason is likely to be the same in both cases. On one hand the *sui generis* semantics that are inevitably associated with identities are difficult to reconcile with the fact that names themselves are rarely unique, such that name tokens don't appear to correspond to generic 'concepts.' On the other, the self same *sui generis* semantics make a discriminative account intuitive because the very fact that, in context, names more often than not succeed in picking out a unique correlated identity means that names are usually unambiguous <u>in context</u>. This means that in many contexts the successful communication of a name will result in the elimination of uncertainty on the part of a hearer (13), as opposed to merely its reduction (which appears to be sufficient for a great many communicative purposes).

---

[14] A previous version of this work was criticized for making "little connection with relevant current work in theoretical pragmatics" and "not much connection with the state of the art in theoretical linguistics, pragmatics, psycholinguistic processing, or children's semantic/pragmatic development," while a reviewer complained that "the specific cases … discussed are all to do with words (names, verbs, nouns, gender systems), while syntax, the key driver of linguistic compositionality, is not mentioned." What I hope is clear to the careful reader (and even future reviewers) is that I hold out no hope that a successful theory of human communication can be built on the idea of 'units of meaning' at any possible level of description, and that as a result, I have little to say about work founded on this idea (the 'state of the art in theoretical linguistics,' 'pragmatics,' 'syntax') other than to note that if the foundations of a scientific theory are wrong, it seems reasonable to assume that its ultimate contributions to human understanding are likely to be minimal.

[15] I acknowledge that many readers will find this account of the communicative function of names unsatisfactory, because will feel that this account fails to satisfy their intuitions about lexical semantics and meaning. I can only reiterate here that what one knows / feels etc., about an identity is entirely independent of how human communicative codes are used to signal them, and note that a large part of the development of a mature science of human communication will involve a re-appraisal of the kinds of phenomena that we can plausibly seek to explain (i.e., it seems likely linguists are no more likely to develop a 'comprehensive theory of meaning' than physicists or horologists are to develop a 'theory of time travel').

(13)    John: *Who wrote that book you told me about?*

Ann: *Richard Feynman*

However, as (14) illustrates, contexts where the successful communication of a name does not result in the elimination of uncertainty on the part of a listener clearly exist.

(14)    John: *Do you like David Bowie?*

Ann: *It depends*

John: *Do you like David Bowie in his Ziggy period?*

Ann: *Yes*

Of course, names are only one, fairly small aspect of human communication. Providing an account of all the other myriad ways in which people communicate using words and sequences of words is beyond the scope of this paper (and its author's current abilities). There are however, many good reasons to believe that accounting for the rest of human communication will be best achieved by an extension of the approach I have outlined here. One obvious reason to support this suggestion is the fact that, as noted above, research has shown compositional accounts of communication to have little going for them other than people appear to be predisposed to intuitively believe them. For most of the past century adherence to these beliefs has led to a situation where the failure to find evidence for compositional theories has *not* been taken as evidence against them. Rather, the absence of evidence has been taken to support claims that people are miraculously able to learn and use all of the various inexplicable aspects of compositional communicative codes in spite of the 'poverty of the stimulus' from which they are supposed to learn them. (Hence the appeal to innate – though ill-defined – stocks of concepts in the quote above.)

*Red Orange Yellow Green Blue Purple Brown Magenta Tan Cyan Olive Maroon Navy Aquamarine Turquoise Silver Lime Teal Indigo Violet Pink Black White Grey/Gray*

**Table 11:** 24 common English color words (from Wikipedia)

To further this line of thought, Table 11 lays out a set of 24 common English color words (taken from wikipedia). Perhaps unsurprisingly given the foregoing, an analysis of the frequency distribution of these words in COCA (plotted in Figure 20) shows that they are geometrically distributed. That is, it seems that when people talk about color in English, they use *white* exponentially more frequently than *black,* which they use exponentially more frequently than *red* which they use exponentially more frequently than *green* which they use exponentially more frequently than *brown* which they use exponentially more frequently than *blue;* and this pattern continues successively for *grey/gray* then *yellow* then *silver* then *orange* then *pink* then *navy* then *olive* then *purple* then *lime* then *tan* then *violet* then *turquoise* then *maroon* then *indigo* then *teal* then *magenta* then *aquamarine* then *cyan*. This pattern is not predicted by compositional theories, but neither does it contradict them. A compositional theorist might suppose that it reflects some bias, or constraint, on people's color concepts, a factor that is independent of communication, and appeal to physiologists or physicists to explain its basis.

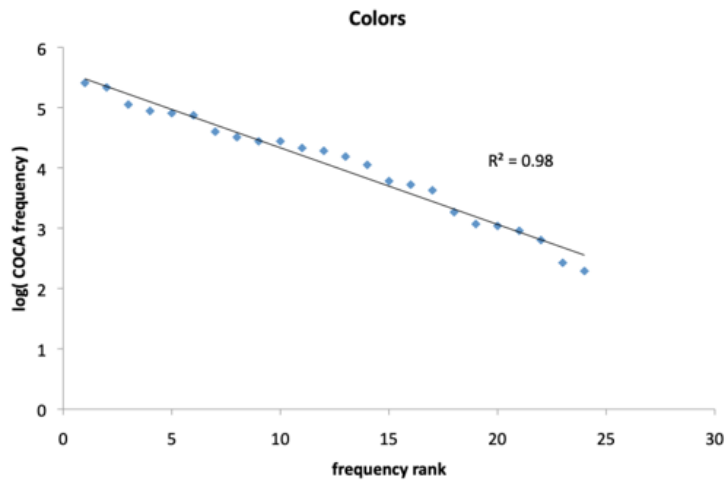

**Figure 20**: Log frequency x frequency rank plot of 24 color English color words ($R^2$=.98).

*Mother, Father, Son, Daughter, Brother, Sister, Uncle, Grandmother, Aunt, Grandfather, Cousin, Grandson, Nephew, Niece, Granddaughter*

**Table 12:** Set of English kinship terms defined by Kemp & Regier (2012).

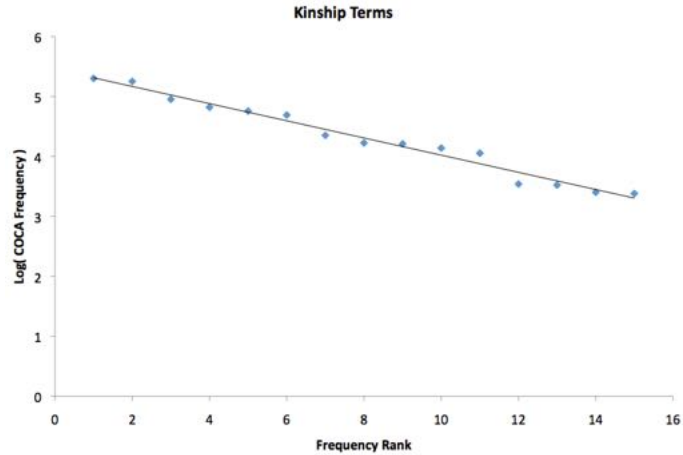

**Figure 21**: Log frequency x frequency rank plot of the English kinship terms defined by Kemp & Regier (2012; $R^2$=.98).

Table 12 then lays out a set of 15 kinship terms taken from Kemp & Regier (2012). An analysis of the frequency distribution of these words in COCA (plotted in Figure 21) shows that they too are geometrically distributed. This pattern is not predicted by compositional theories either, but once again, neither does it contradict them. Our compositional theorist might suppose that it too reflects some bias, or constraint, this time on people's kinship concepts, a factor that is also independent of communication, and appeal to sociologists or anthropologists to explain its basis (again, contrary to the claim noted above, Chomsky, 2000, it seems that the empirical facts always appear to leave open many possibilities).

By contrast, a key piece of evidence I offered in support of the idea that names support a discriminative communicative function is the way they are distributed (and from the apparent universality of these distributions). Section 5 further showed that even if one were to object to my seeking to abstract my information theoretic account of communication from names to the rest of human communication by arguing that names are somehow 'special,' it turns out that many other 'more usual' communicative concepts (such as those associated with nouns and verbs) have the same distributional structure. Moreover, as Figure 22 appears to indicate, many of the 'biases and constraints' that shape people's use of words across different domains appear to result suspiciously similar patterns. These patterns make perfect sense from a discriminative, information theoretic, perspective, but once again, while they are not at odds with compositional accounts, neither are they predicted by the various intuitions that appear to drive people's implacable faith in them.

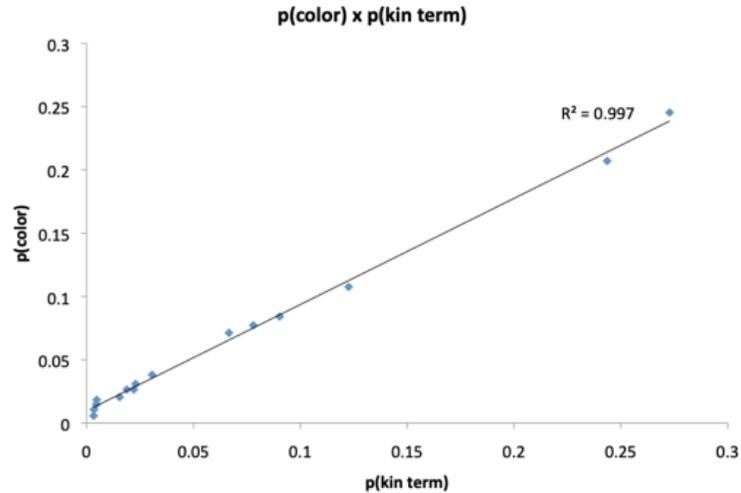

**Figure 22**: Point-wise comparison of the rank probabilities of English color words (calculated over a set of 24 items) and English kin terms (calculated over a set of 15 items; $R^2=.997$).

Thus while it is true that with sufficient time and effort, one could come up with plausible-sounding post hoc stories about biases, constraints etc., that account for all of the skewed variance in the frequencies of these supposedly compositional items (variance that compositional intuitions utterly fail to predict), at some point it seems that a duck test[16] might be in order. If natural languages are distributed the way that deductive, non-compositional communicative codes are distributed, if the lexical contrasts within domains maximize discrimination in the way that the codewords in a communicative code maximize discrimination, and if the information in human codes increases according to communicative demands in the way one would expect the information in a communicative code to increase as its coverage increases, this is probably because natural languages are in fact deductive, non-compositional communication systems.

*8.2 Information and meaning*

Since a great deal of the last part of this paper has been devoted to explaining to readers why compositional accounts of human communication – or, as these theories usually describe it, 'language' – are a theoretical dead end, I feel that it behooves me to end in a more upbeat manner by providing furthers example of how information theoretical accounts of communication can positively increase our understanding of meaningful questions. To this end, in closing I will briefly consider how the approach I have described here can be used to illuminate what is often called the "Easterlin paradox" in the relationship between wealth and human happiness. In doing so, I will try to highlight a few outstanding questions raised by this approach, as well as the benefits that information theory brings to this specific question.

---

[16] If it looks like a duck, swims like a duck, and quacks like a duck, it probably is a duck.

The "Easterlin paradox" describes a curious finding in the relationship between wealth and human happiness. Put simply, it has been argued that although at any given point in time a relationship between wealth and happiness can be seen (such that, say, if richer and poorer countries / citizens are compared at a point in time, life satisfaction increases with the absolute amount of GDP per capita), across time there seems to be no significant relationship between the rate of improvement in happiness and the rate of economic growth (see e.g., Easterlin & Angelescu, 2009; Easterlin, 2013; Hills, Proto & Sgro, 2015). In this work, happiness is measured by overall life satisfaction, which is operationalized in terms of people's responses to the following question (Easterlin & Angelescu, 2009):

(15) *"All things considered, how satisfied are you with your life as a whole these days"*

Accordingly, it follows that the happiness / income paradox is more accurately put as follows: while at any point in time the degree to which people report being satisfied increases as income increases, it turns out that over time the degree to which people report being satisfied does not increase as incomes increase.

How can an information theoretical account of communication increase our understanding of this matter? To begin with, it is important to note this paradox arises under (relies on) a critical assumption, namely that when people are asked, *how satisfied are you with your life*, the meaning of *satisfied* is constant over time. This, of course, is a standard compositional assumption: *satisfied* is assumed to be associated with a 'concept,' and a further (implicit) assumption is that concepts are generally stable over time, except when they are not; at which point compositional theories usually don't have much to say about how concepts change, because compositional theories general analyses concepts in isolation (Ramscar & Port, 2015). By contrast, as I have emphasized throughout the foregoing (and as figure 12 in particular serves to illustrate) information is a property of systems, and these systems balance the constraints imposed by the requirement to communicate specific messages (e.g., about identities), with the need to make those specific messages informative (which also embraces predictable and learnable) in a community that needs a number of different specific messages to be communicable.

This view of communication is, as I have repeatedly stressed, concerned with uncertainty reduction. It follows therefore that over time, people might be expected to use specific words in some kind of relation to the degree to which they relate to the way uncertainty is distributed in the environment. Thus, for example, in a series of analyses (Ramscar, 2015) of *old-* word sequences such as *old man*, *old woman,* I found that the frequency of *old-* word sequences declined as the number of elderly people in the population increased across the late 20[th] Century. While this might seem paradoxical from a compositional perspective (where on might expect that the presence of more old men might predict more talk of *old men*), it is entirely consistent with the account of names presented earlier, since it follows that the more old men there are, the less informative (discriminative) talk of *old men* will become in context.

Since human communication can in many cases be seen as the utilization of a code in order to reduce uncertainty,[17] it follows that when changes in experiential context change the informativity of an aspect of the code, people's use of that aspect of the code is likely to change. It also follows that its informativity within the code, both in context and perhaps across contexts, will change. In this case, the information that *old* contributes to a listener's expectation that *man, men, woman* or *women* will occur in spoken contexts can be shown to have changed considerably in the past 50 years. It follows also that the meaning of *old man, old men, old woman* and *old women* are likely to have changed as well. While I have no idea how one might hope to quantify what these changes in meaning are (or even if this question is well-posed, Wittgenstein, 1953), an information theoretic approach to communication does at least offer methods for evaluating whether we might expect changes to have taken place.

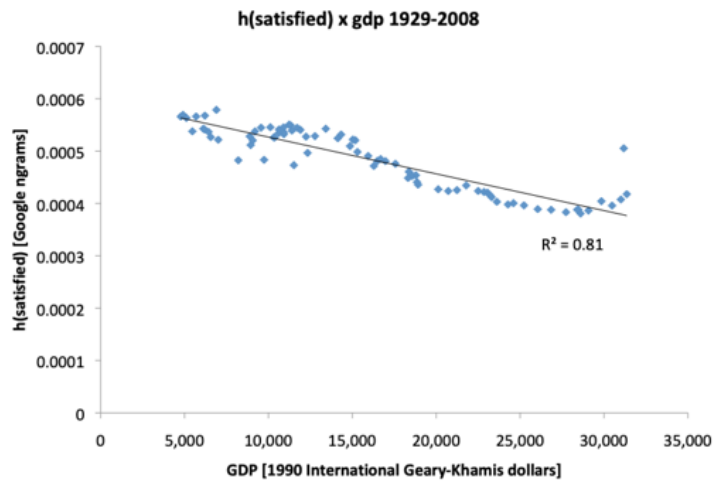

**Figure 23**: the entropy of the word *satisfied* in Google ngrams from 1929-2008, plotted against US per capita GDP[18] over the same period.

---

[17] This is not to deny that non-verbal and non-lexical codes also clearly exist, or that they contribute enormously to human communication.

[18] Measured in 1990 International Geary-Khamis dollars. Data from: www.ggdc.net/maddison/historical_statistics/

With this in mind, and returning to the happiness – wealth paradox, I analyzed the informativity of the word *satisfied* (operationalized as its entropy in Google ngrams) in relation to US per capita GDP over the period 1929-2008 (Figure 23). As can be seen, it is clear that across this period, increases in wealth are strongly associated with changes in the informativity of *satisfied* ($R^2$=.81).

Of course, it might be objected that since my analyses of Delaware names, *old men* etc. and *satisfied* all reveal declines in the informativity of specific words across time that this might just be a general property of the code: perhaps all words decline in informativity. To at least attempt to control for this possibility, I analyzed the informativity of the word *poverty* (operationalized as its entropy in Google ngrams) in relation to US per capita GDP over the period 1929-2008. Since I argued that the increases in general satisfaction that result from increases in GDP might lead people to talk less about being *satisfied* (because *satisfied* will become less informative when more people are satisfied), it follows that the opposite ought to be true of *poverty*. As levels of poverty decrease, then the word poverty ought to increase in its informativity (*poverty* will be uninformative when everyone is in poverty, and its informativity will increase as poverty becomes less widespread).

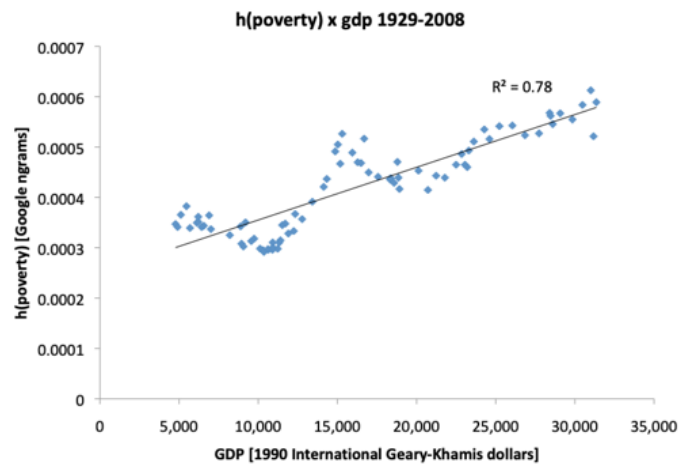

**Figure 23**: the entropy of the word *poverty* in Google ngrams from 1929-2008, plotted against US per capita GDP over the same period.

As figure 23 shows, whereas increases in wealth were strongly associated with a decline in the use of *satisfied* between 1929 and 2008, the opposite was true of *poverty* – as America's wealth increased, talk about *poverty* amongst American increased as well ($R^2$=.78). In line with my remarks above, it is worth highlighting two things that these results seem to indicate: first, as people have become wealthier, they talk about being *satisfied* less, presumably because, if we suppose *satisfied* means 'satisfied,' the fact that more and more people are satisfied makes talk about being satisfied less meaningful; and second, because people talk about being *satisfied* less as they become wealthier, it follows that talk about being *satisfied* must increasingly occur in fewer, more specialized contexts, such that if we accept that the use of *satisfied* contributed something to the meaning of 'satisfied' in 1929, then the meaning of 'satisfied' must have changed by 2008. [19] (And of course, when it comes to *poverty*, the opposite pattern seems to hold.)

Interestingly, an analysis of the informativity of the word *happy* (operationalized as its entropy in Google ngrams) in relation to US per capita GDP over the period 1929-2008 appears to confirm the wealth-happiness paradox, since it revealed the same U-shaped relationship between wealth and poverty reported by Easterlin and colleagues (Figure 24).

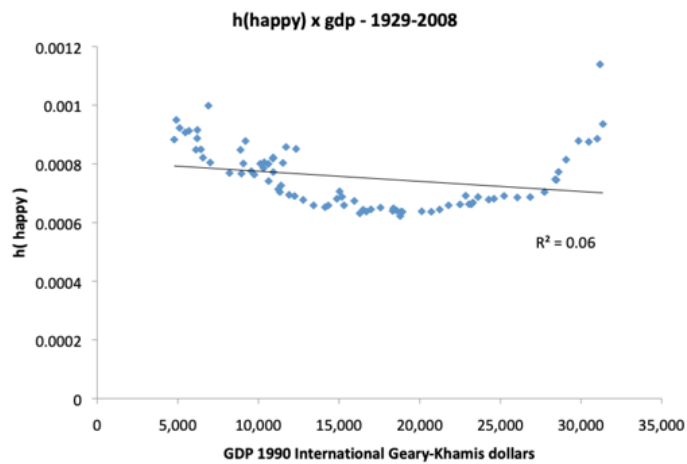

**Figure 24**: the entropy of the word *happy* in Google ngrams from 1929-2008, plotted against US per capita GDP over the same period.

This raises some interesting questions, an obvious one being whether the tendency in the literature to equate people's *being satisfied* with *happiness* in comparisons of wealth and attitude is justified, given that, in terms of informativity at least, *satisfied* and *happy* interact very differently with wealth over time. Among the other obvious questions it raises are whether these correlations actually mean anything, and if they do, what, exactly are they supposed to be telling us? I will not attempt to answer these questions here, not least because I suspect that each of them represents a serious research topic in its own right. But I will try to briefly sketch out why I feel they are useful questions, and why I think an information theoretic approach can shed useful light on them.

---

[19] This paragraph clearly highlights the perils involved in using words to talk about word meanings.

First, what I hope is clear is that while simply assuming that words like *satisfied, poverty* and *happy* have fixed associated concepts will serve to affirm the paradox and satisfy people's compositional intuitions, it will fail to explain the systematic changes observed in the use of the words *satisfied* and *poverty* over time, which, if one cares about relations between wealth and well-being, seem worth exploring. Second, while it is true that the comparison between the informativity of *happy* and gdp observed in figure 24 supports the wealth – happiness paradox, the fact that this analysis revealed the same U-shaped relationship between wealth and poverty reported by Easterlin and colleagues supports the overall suggestion here that informativity may be a useful way of looking at meaning over time, albeit that this matter is muddied considerably by the fact that data reporting attitudes analyzed by Easterlin and colleagues were responses to satisfaction questions. This raises still more questions, not the least of which is whether the answers that people provide in response to survey questions answer the questions that researchers want to ask (Bertrand & Mullainathan, 2001).

Which is to say, finally, that if one cares about things like relations between wealth and well-being, then if people's ideas of wealth and well-being are at all influenced by the way words like *wealth and well-being* are used in the codes we use to communicate them, improving our scientific understanding of these codes can only help. In this regard, it is interesting to consider the degree to which we might expect the words *happy, satisfied* and *poverty* to be directly related to observable economic data in the first place. To take the last of these first, if we assume that *poverty* is typically used to talk about existing in a state where ones day to day basic living needs are not met (i.e., where one daily faces some kind of existential threat), then it seems reasonable to expect that the informativity of the word *poverty* might be expected to change as general living standards improve. Similarly, if we assume that people tend to be *satisfied* when their day to day basic living *are* met (i.e., where they are not faced with any kind of existential threat), then it also seems reasonable to expect that the informativity of the word *satisfied* might be expected to change as general living standards improve.

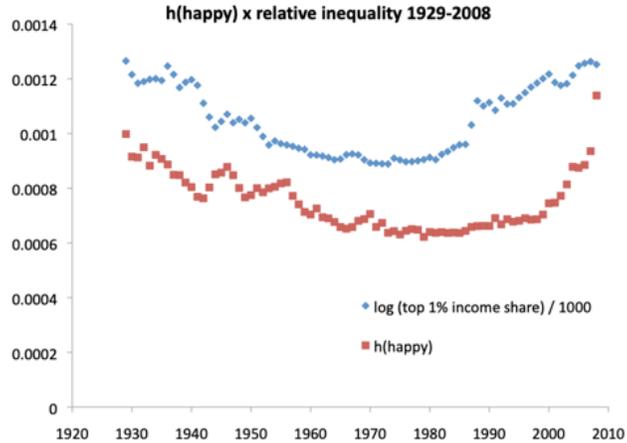

**Figure 25**: the entropy of the word *happy* in Google ngrams from 1929-2008, plotted against the scaled log of the income share of the top 1% of earners[20] over the same period.

This then leaves us with *happy,* which seems to have a rather more complex relationship to wealth than *poverty* or *satisfied,* not least because, for example, getting a 20% pay rise might alleviate the poverty of a low paid worker (and increase their general satisfaction), it might not necessarily make them happy, especially if they learn that all of their colleagues are getting, say, an 80% pay rise, an analysis that seems supported by relationship between the entropy of *happy* and a common measure of relative income inequality (Figure 25; $R^2=.47$). It is worth stressing again that I do not want to claim that these analyses are "right." That isn't the point of this exercise. Rather what I hope is obvious is that addressing these questions from a perspective that treats words as parts of information systems can cast an interesting and potentially helpful light on them. It can help illuminate some important factors that one might to control for if one does care about things like relations between wealth and well being, as many governments seem to do, and it raises – and allows one to even explore objectively – a very interesting question: if one wants to get at the answer to a complex question like the relationship between wealth and well being, is it better to ask people what they think (at a given point in time and a particular context), or to look at the way that people's communicative behavior reflects what they think across time, and across contexts.

---

[20] Data from https://eml.berkeley.edu/~saez/TabFig2012prel.xls

## 9. Conclusion

I have described an account of how human communication works based on well-established theories of learning[21] and communication, and have used the predictive power of this account to uncover and describe the way in which the cultural environment has evolved a set of remarkable structures to support human communication and the learning of human communicative skills; structures that were hitherto undiscovered. Finally, I have described how although these structures are largely incompatible with – and unpredicted by – contemporary linguistic theory, they are entirely consistent with what we know about formal theories of communication, learning and coding (Shannon, 1949; Ramscar et al, 2010). Given this last point, perhaps the most surprising aspect of these data is just how surprising most students of language and communication will find them to be.

---

[21] Across species learning has evolved in a way that enables individuals to respond in ways that are highly sensitive to information in the environment (Rescorla, 1988). Meanwhile, the development of cognitive control / selective attention (which allows individuals to self-direct their learning) develops very slowly in humans, such that young humans appear to be particularly pre-disposed to the learning of the conventions that communications systems appear to rely on (Ramscar & Gitcho, 2007). Taken together with the data described here, these considerations indicate that interactions between linguistic behavior, learning and its development within communities are capable over time of producing self-organizing communication systems, as well as the remarkable statistical – and ontological – structures described here.